\documentclass{article} % For LaTeX2e

\usepackage{iclr2024_conference}

\usepackage{hyperref}
\usepackage{url}
\usepackage[utf8]{inputenc} % allow utf-8 input
\usepackage[T1]{fontenc}    % use 8-bit T1 fonts
\usepackage{url}            % simple URL typesetting
\usepackage{booktabs}       % professional-quality tables
\usepackage{amsfonts}       % blackboard math symbols
\usepackage{nicefrac}       % compact symbols for 1/2, etc.
\usepackage{microtype}      % microtypography
\usepackage{xcolor}         % colors
\usepackage[final]{changes}
\usepackage{amssymb}
\usepackage{amsmath}
\usepackage{tabularx}
\usepackage{enumitem}
\usepackage{multirow}
\newcolumntype{Y}{>{\centering\arraybackslash}X}
\newcommand{\sref}[1]{\textsuperscript{\ref{#1}}}
\definechangesauthor[name={Gabina}, color=purple]{GS}
\definechangesauthor[name={Michal}, color=orange]{MV}
\definechangesauthor[name={Dusan}, color=cyan]{DF}
\definechangesauthor[name={Karla}, color=green]{KS}

\newcommand{\dataset}{CdSprites+\ }
\usepackage{pifont}
\newcommand\mycomment{}
\newcommand{\cmark}{\ding{51}} % Checkmark
\newcommand{\xmark}{\ding{55}} % Cross

\title{Benchmarking Multimodal Variational Autoencoders: \dataset Dataset and Toolkit}

\author{%
  Gabriela Sejnova$^{1}$  \; \; \;  Michal Vavrecka$^{1}$  \; \; \;  Karla Stepanova$^{1}$\\
  $^1$Czech Institute of Informatics, Robotics and Cybernetics\\
  Czech Technical University in Prague,
  Prague, Czech Republic\\
  \texttt{gabriela.sejnova@cvut.cz} \\
  \And
 Tadahiro Taniguchi$^{2,3}$ \\
 $^2$Graduate School of Informatics, Kyoto University \\
 Yoshida-Honmachi, Sakyo, Kyoto 606-8501, Japan \\
 \texttt{taniguchi@i.kyoto-u.ac.jp} \\
 $^3$Research Organization of Science and Technology,
Ritsumeikan University \\
        1-1-1 Noji-Higashi, Kusatsu, Shiga 525-8577, Japan
}
\begin{document}
'\nolinenumbers

\maketitle

\begin{abstract}
Multimodal Variational Autoencoders (VAEs) have been the subject of intense research in the past years as they can integrate multiple modalities into a joint representation and can thus serve as a promising tool for both data classification and generation. Several approaches toward multimodal VAE learning have been proposed so far, their comparison and evaluation have however been rather inconsistent. One reason is that the models differ at the implementation level, another problem is that the datasets commonly used in these cases were not initially designed to evaluate multimodal generative models. This paper addresses both mentioned issues. First, we propose a toolkit for systematic multimodal VAE training and comparison. The toolkit currently comprises 4 existing multimodal VAEs and 6 commonly used benchmark datasets along with instructions on how to easily add a new model or a dataset. Second, we present a disentangled bimodal dataset designed to comprehensively evaluate the joint generation and cross-generation capabilities across multiple difficulty levels. We demonstrate the utility of our dataset by comparing the implemented state-of-the-art models.
\end{abstract}

\section{Introduction}
%- Needs to be shortened a lot. Main message: \\ - Many new VAE models, comparison inconsistent with regards to datasets, implementations, used metrices - add references to evaluation papers \\ - no currently available tool that would be model- and dataset-agnostic \\ - datasets that are used are either same modalities or different modalities but with workarounds such as nearest neighbour matching of image features. Also real world images and text annotations are hard to objectively evaluate (CUB).\\
%- our first contribution is the toolkit which enables systematic comparison, hyperparameter tuning and simple extension for new datasets, models etc. \\
%- we provide dataset that is clear to evaluate, fast to generate large number of samples and the cross generation can be automatically evaluated in the image-to-text translation. \\
%- we want to establish a unified benchmark for further multimodal vae development so that it is reproducible but also replicable\\
%}

Variational Autoencoders (VAEs) \citep{kingma2013auto} have become a multipurpose tool applied to various machine perception tasks and robotic applications over the past years
\citep{pu2016variational}\citep{xu2017variational}\citep{nair2018visual}. For example, some of the recent implementations address areas such as visual question answering \citep{chen2019multi}, visual question generation \citep{uppal2021c3vqg} or emotion recognition \citep{yang2019attribute}. Recently, VAEs were also extended for the integration of multiple modalities, enabling mapping different kinds of inputs into a joint latent space. It is then possible to reconstruct one modality from another or to generate semantically matching pairs, provided the model successfully learns the semantic overlap among them. 

Several different methods for joint multimodal learning have been proposed so far and new models are still being developed \citep{wu2018multimodal}\citep{shi2019variational}\citep{sutter2021generalized}.
Two of the widely recognized and compared approaches are the MVAE model \citep{wu2018multimodal}, utilizing the Product of Experts (PoE) model to learn a single joint distribution of the joint posterior, and MMVAE \citep{shi2019variational}, using a Mixture of Experts (MoE). \cite{sutter2021generalized} also recently proposed a combination of these two architectures referred to as Mixture-of-Products-of-Experts (MoPoE-VAE), which approximates the joint posterior on all subsets of the modalities using a generalized multimodal ELBO.\looseness=-1

The versatility of these models (i.e. the possibility to classify, reconstruct and jointly generate data using a single model) naturally raises a dispute on how to assess their generative qualities. \cite{wu2018multimodal} used test set log-likelihoods to report their results, \cite{shi2019variational} proposed four criteria that measure the coherence, synergy and latent factorization of the models using various qualitative and quantitative (usually dataset-dependent) metrics. Evaluation for these criteria was originally performed on several multimodal benchmark datasets such as MNIST \citep{deng2012mnist}, CelebA \citep{liu2015faceattributes}, MNIST and SVHN combination \citep{shi2019variational} or the Caltech-UCSD Birds (CUB) dataset \citep{wah2011caltech}. All of the mentioned datasets comprise images paired either with labels (MNIST, CelebA), other images (MNIST-SVHN, PolyMNIST) or text (CelebA, CUB). Since none of these datasets was designed specifically for the evaluation of multimodal integration and semantic coherence of the generated samples, their usage is in certain aspects limited. %\deleted{More specifically, the image-label datasets are very distant from the potential real-world applications of multimodal VAEs, the image-image datasets currently only comprise digits with varying amounts of noise (such as MNIST-SVHN or PolyMNIST) and the modalities are thus conceptually identical. In comparison, the image-text datasets (CUB, CelebA) are far more challenging for the state-of-the-art models as they comprise real-world, manually annotated data. For this reason, their qualitative evaluation is problematic due to a high amount of noise and subjectively assessed attributes (e.g. an  \textit{attractive and young face} in the case of CelebA).}
More specifically, datasets comprising real-world images/captions (such as CUB or CelebA) are highly noisy, often biased and do not enable automated evaluation. This makes these datasets unsuitable for detailed analysis of the model's individual capabilities such as joint generation, cross-generation or disentanglement of the latent space. On the other hand, simpler datasets like MNIST and SVHN do not offer different levels of complexity, provide only a few categories (10-digit labels) and cannot be used for generalization experiments. %Moreover, it has been shown that the current multimodal VAEs fail to reconstruct these images and their features need to be used instead, } \citep{shi2019variational}, \citep{daunhawer2021limitations} 

Due to the above-mentioned limitations of the currently used benchmark datasets and also due to a high number of various implementations, objective functions and hyperparameters that are used for the newly developed multimodal VAEs, the conclusions on which models outperform the others substantially differ among the authors \citep{shi2019variational}, \citep{kutuzova2021multimodal}, \citep{daunhawer2021limitations}. Recently, \cite{daunhawer2021limitations} published a comparative study of multimodal VAEs where they conclude that new benchmarks and a more systematic approach to their evaluation are needed. 

In this paper, we propose a unification of the various implementations into a single toolkit which enables training, evaluation and comparison of the state-of-the-art models and also faster prototyping of new methods due to its modularity. The toolkit is written in Python and enables the user to quickly train and test their model on arbitrary data with an automatic hyperparameter grid search. By default, the toolkit includes 6 currently used datasets that have been previously used to compare these models.  Moreover, we include our new synthetic dataset that enables a more systematic evaluation called \dataset (\textit{Captioned disentangled Sprites +}) and is built on top of the existing dSprites (disentangled Sprites) dataset used to assess the capabilities of unimodal VAEs \citep{dsprites17}. While dSprites comprises grayscale images of 3 geometric shapes of various sizes, positions and orientations, \dataset also includes natural language captions as a second modality and 5 different difficulty levels based on the varying number of features (i.e. varying shapes for Level 1 and varying shapes, sizes, colours, image quadrants and backgrounds in Level 5). \dataset enables fast data generation, easy evaluation and also a gradual increase of complexity to better estimate the progress of the tested models. It is described in greater detail in Section \ref{datasetdesc}. For a brief comparison of the benchmark dataset qualities, see also Table \ref{tab:datasettable}.%Moreover, we design a synthetic bimodal dataset which is incorporated into this toolkit and enables automated quantitative and qualitative evaluation. The dataset comprises images of geometric shapes and their textual descriptions and offers several complexity levels for a detailed assessment of the models' capabilities. We also demonstrate the utility of the toolkit and dataset on a set of experiments.

In conclusion, the contributions of this paper are following:

\begin{enumerate}
 \item We propose a public toolkit which enables systematic development, training and evaluation of the state-of-the-art multimodal VAEs and their comparison on the commonly used multimodal datasets.
  \item We provide a synthetic image-text dataset called \dataset designed specifically for the evaluation of the generative capabilities of multimodal VAEs on 5 levels of complexity.
\end{enumerate}

The toolkit and code for the generation of the dataset (as well as a download link for a ready-to-use version of the dataset) are available on GitHub \footnote{\url{www.github.com/gabinsane/multimodal-vae-comparison}\label{git}}.

\section{Related Work}

In this section, we first briefly describe the state-of-the-art multimodal variational autoencoders and how they are evaluated, then we focus on datasets that have been used to demonstrate the models' capabilities.

\subsection{Multimodal VAEs and Evaluation}
\label{ssec:rweval}

Multimodal VAEs are an extension of the standard Variational Autoencoder (as proposed by \cite{kingma2013auto}) that enables joint integration and reconstruction of two or more modalities.  During the past years, a number of approaches toward multimodal integration have been presented  \citep{suzuki2016joint}, \citep{wu2018multimodal}, \citep{shi2019variational}, \citep{vasco2020mhvae}, \citep{sutter2021generalized}, \citep{joy2021learning}. For example, the model proposed by \cite{suzuki2016joint} learns the joint multimodal probability through a joint inference network and instantiates an additional inference network for each subset of modalities. % which is computationally costly in the case of more than two modalities.
A more scalable solution is the MVAE model \citep{wu2018multimodal}, where the joint posterior distribution is approximated using the product of experts (PoE), exploiting the fact that a product of individual Gaussians is itself a Gaussian. %One drawback here is the possible imbalanced precision of individual experts which can cause a shift of the predicted joint distribution density toward the more precise expert. 
In contrast, the MMVAE approach \citep{shi2019variational} uses a mixture of experts (MoE) to estimate the joint variational posterior based on individual unimodal posteriors. %Here the probability density is shared among all the experts, avoiding the problem with imbalanced modalities as in PoE. 
The MoPoE architecture from \cite{sutter2021generalized} combines the benefits of PoE and MoE approaches by computing the joint posterior for all subsets of modalities. %, providing a valid bound on the joint log-likelihood (which is not guaranteed in MVAE) while preserving the ability to aggregate multiple modalities (contrary to MMVAE). 
Another recent extension is the DMVAE model, in which the authors enable the encoders to separate the shared and modality-specific features in the latent space for a disentangled multimodal VAE \citep{lee2021private}.
\added{While there are also other recently proposed multimodal VAEs  (\cite{sutter2020multimodal}, \cite{daunhawer2021self}, \cite{liang2022foundations}, \cite{palumbo2023mmvae+}), in this paper, we highlight the 4 abovementioned models that can be considered representative of the individual approaches as summarized by \citet{suzuki2022survey}.  For example, MMVAE does not learn the joint posterior distribution (while the 3 remaining models do), MoPoE is scalable to multiple modalities only under very high computational costs (compared to the other models) and DMVAE is the only model that learns modality-specific (private) latent variables.}

The evaluation of the above-mentioned models has also evolved over time. \cite{wu2018multimodal} measured the test marginal, joint and conditional log-likelihoods together with the variance of log importance weights. \cite{shi2019variational} proposed four criteria for evaluation of the generative capabilities of multimodal VAEs: \textit{coherent joint generation}, %i.e. the amount of semantic coherence among samples of different modalities stemming from the same latent value, 
\textit{coherent cross-generation}, %i.e. generation of semantically consistent samples in one modality conditioned on another modality,
 \textit{latent factorisation} %of the modalities into their shared and private subspaces, 
 and \textit{synergy}. %, i.e. improvement of the generative quality for individual modalities after their joint mapping with one another. 
 All criteria are evaluated both qualitatively (through empirical observation of the generated samples) and quantitatively: by adopting pre-trained classifiers for evaluation of the generated content, by training a classifier on the latent vectors to test whether the classes are distinguishable in the latent space, or by calculating the correlation between the jointly and cross-generated samples using the Canonical Correlation Analysis (CCA). 

Besides certain dataset-dependent alternatives, %(such as the Fréchet inception distance used for image quality evaluation by Daunhawer et al. \citep{daunhawer2021limitations})
the most recent papers use a combination of the above-mentioned metrics for multimodal VAE comparison \citep{sutter2021generalized},  \citep{joy2021learning} \citep{daunhawer2021limitations}, \citep{kutuzova2021multimodal}. Despite that, the conclusions on which model performs the best according to these criteria substantially differ. According to a thorough comparative study from \cite{daunhawer2021limitations}, none of the current multimodal VAEs sufficiently fulfils all of the four criteria specified by \cite{shi2019variational}. % and they also fail on more complex datasets such as CUB \citep{wah2011caltech}. 
 Furthermore, the optima of certain training hyperparameters might be different for each model (as was proven e.g. with the regularisation parameter $\beta$~ \citep{daunhawer2021limitations}), which naturally raises the need for automated and systematic comparison of these models over a large number of hyperparameters, datasets and training schemes. \cite{sutter2021generalized} released public code which allows comparison of the MoE, PoE, MoPoE approaches - however, the experiments are dataset-dependent and an extension for other data types would thus require writing a substantial amount of new code. 

In this paper, we propose a publicly available toolkit for systematic training, evaluation and comparison of the state-of-the-art multimodal VAEs. Special attention is paid to hyperparameter grid search %(we provide automated scripts for the generation of training configs with desired parameter variations)
and automatic visualizations of the learning during training. To our knowledge, it is the only model- and dataset-agnostic tool available in this area that would allow fast implementation of new approaches and their testing on arbitrary types of data.

\subsection{Multimodal Datasets}
\label{ssec:rwdatasets}

At this time, there are several benchmark datasets commonly used for multimodal VAE evaluation. The majority is bimodal, where one or both modalities are images. In some of the datasets, the bimodality is achieved by splitting the data into images as one modality and the class labels as the second - this simplifies the evaluation during inference, %\deleted{ yet such dataset is far from real-world use cases (e.g. image-text translation) where both modalities can be complex and high-dimensional}
 yet such dataset does not enable e.g. testing the models for generalization capabilities and the overall number of classes is usually very low. Examples of such datasets are MNIST \citep{deng2012mnist}, FashionMNIST \citep{xiao2017fashion} or MultiMNIST \citep{sabour2017dynamic}. Another class are image-image datasets such as MNIST and SVHN \citep{netzer2011reading} (as used in \cite{shi2019variational}), where the content is semantically identical and the only difference is the form (i.e. handwritten digits vs. house numbers). This can be seen as integrating two identical modalities with different amounts of noise - while such task might be relevant for some applications, %\deleted{it is again distant from practical use cases.}
it does not evaluate whether the model can integrate multiple modalities with completely different data representations.

An example of a bimodal dataset with an image-text combination is the CelebA dataset containing real-world images of celebrities and textual descriptions of up to 40 binary attributes (such as \textit{male, smiling, beard} etc.) \citep{liu2015faceattributes}. While such a dataset is far more complex, the qualitative evaluation of the generated samples is difficult and ambiguous due to the subjectiveness of certain attributes (\textit{attractive, young, wearing lipstick}) combined with the usually blurry output images generated by the models. Another recently used image-text dataset is the Caltech-UCSD Birds (CUB) dataset \citep{wah2011caltech} comprising real-world bird images accompanied with manually annotated text descriptions (e.g. \textit{this bird is all black and has a long pointy beak}). %\deleted{One problem here is again the fact that such descriptions can be subjective and in some cases, the attributes cannot even be observed in the small images. Another drawback is that the}
However, these images are too complex to be generated by the state-of-the-art models (as proven by \cite{daunhawer2021limitations}) and the authors thus only use their features and perform the nearest-neighbour lookup to match them with an actual image \citep{shi2019variational}. This also makes it impossible to test the models for generalization capabilities (i.e., making an image of a previously unseen bird).  

Besides the abovementioned datasets that have already been used for multimodal VAE comparison, there are also other multimodal (mainly image-text) datasets available such as Microsoft COCO \citep{lin2014microsoft} or Conceptual Captions \citep{sharma2018conceptual}. Similar to CUB, these datasets include real-world images with human-made annotations, which can be used to estimate how the model would perform on real-world data. However, they cannot be used for deeper analysis and comparison of the models as they are very noisy (including typos, wrong grammar, many synonyms, etc.), subjective, biased or they require common sense (e.g., “piercing eyes”, “clouds in the sky” or “powdered with sugar”). This makes the automation of the semantic evaluation of the generated outputs difficult. %CLEVR dataset \citep{johnson2017clevr}, comprising synthetic images and natural language questions. However, these questions require complex logical reasoning which is currently out of bounds for the state-of-the-art multimodal VAEs. \added{An adaptation suited for multimodal VAEs is 
\added{A more suitable example would be the Multimodal3DIdent dataset \citep{daunhawer2023identifiability}, comprising synthetic images with textual descriptions adapted from CLEVR \citep{johnson2017clevr}. However, this dataset does not have distinct levels of difficulty that would enable to distinguish what the models can learn and what is too challenging.}

In conclusion, the currently used benchmark datasets for multimodal VAE evaluation are not optimally suited for benchmarking due to oversimplification of the multimodal scenario (such as the image-label combination or image-image translation where both images show digits), or, in the opposite case, due to overly complex modalities that are challenging to reconstruct and difficult to evaluate. %We believe that the optimal dataset would be such that resembles real-world applications (i.e. includes two or more different and high-dimensional modalities), is moderately challenging (i.e. the models can reconstruct the raw data) and offers clear qualitative and quantitative evaluation. 
In this paper, we address the lack of suitable benchmark datasets for multimodal VAE evaluation by proposing a custom synthetic dataset called \dataset (\textit{Captioned disentangled Sprites +}). This dataset extends the unimodal dSprites dataset \citep{dsprites17} with natural language captions, additional features (such as colours and textures) and 5 different difficulty levels. It is designed for fast data generation, easy evaluation and also for the gradual increase of complexity to better estimate the progress of the tested models. It is described in greater detail in Section \ref{datasetdesc}. For a brief comparison of the benchmark dataset qualities, see also Table \ref{tab:datasettable}.

\section{\dataset Dataset}
\label{datasetdesc}

\begin{figure}[!htp]
\centering
\includegraphics[width=.9\textwidth]{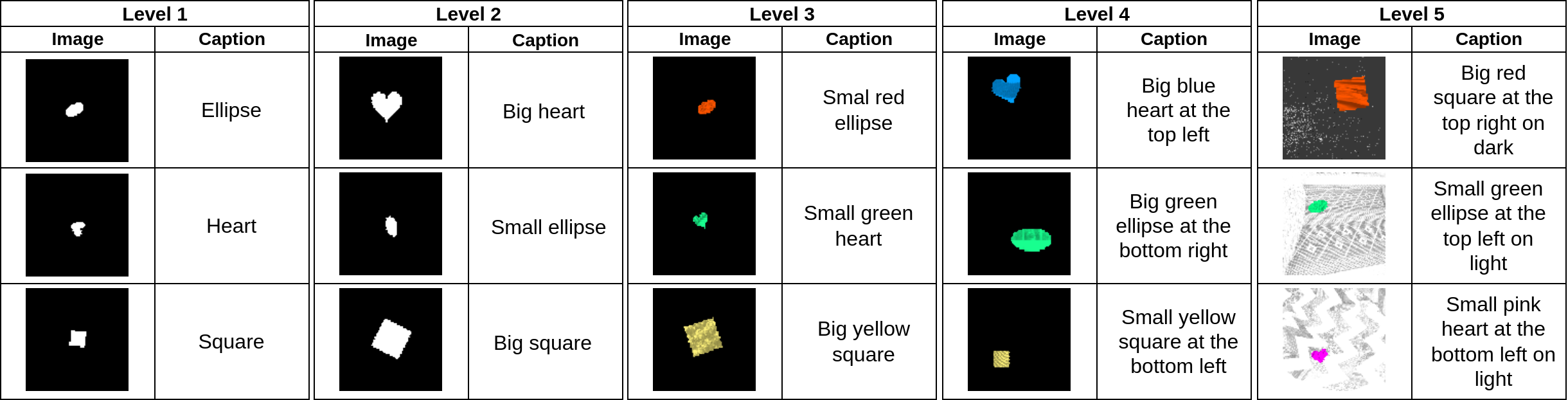}
\caption{Examples of our proposed \dataset dataset. The dataset contains RGB images (left columns) and their textual descriptions (right columns). We provide 5 levels of difficulty (\textit{left} to \textit{right}).  Level 1 only varies the shape attribute, Level 2 varies shape and size, Level 3 varies also the colour attribute, Level 4 varies the position and Level 5 varies also the background shade. See a more detailed description of the dataset in Section \ref{datasetdesc}.}
\label{fig:data}
\end{figure}

We propose a synthetic image-text dataset called \dataset (\textit{Captioned disentangled Sprites +}) for a clear and systematic evaluation of the multimodal VAE models. The main highlights of the dataset are its scaled complexity, quick adjustability towards noise or the number of classes and its rigid structure which enables automated qualitative and quantitative evaluation.  This dataset extends the unimodal dSprites dataset \citep{dsprites17} with natural language captions, additional features (such as colours and textures) and 5 different difficulty levels. A ready-to-use version of the dataset can be downloaded on the link provided in our repository  \textsuperscript{\ref{git}}, the users can also modify and generate the data on their own using the code provided in our toolkit.

\subsection{Dataset structure}
The dataset comprises images of geometric shapes (64x64x3 pixels) with a defined set of 1 - 5 attributes (designed for a gradual increase of complexity) and their textual descriptions. The full variability covers 3 shape primitives (heart, square, ellipse), 2 sizes (big, small), 5 colours, 4 locations (top/bottom + left/right) and 2 backgrounds (dark/light), creating 240 possible unique feature combinations (see Fig. \ref{fig:data} for examples and Table \ref{tab:stats} in the Appendix for the statistics and Section \ref{sec:cdsprites} for further information). To avoid overfitting, the colours of the shapes are textured as well as the changing backgrounds in level 5. Another source of noise (besides textures) is the randomised positions and orientations of the shapes. The default version of the dataset consists of \mycomment{75k (Level 1) 120k (Level 2), 300k (Level 3), 600k (Level 4) and 960k (Level 5),} samples (where 10~\% is used for validation).

The textual descriptions comprise 1 - 8 words (based on the selected number of attributes) which have a rigid order within the sentence (i.e. size, colour, shape, position and background colour). To make this modality challenging to learn, we do not represent the text at the word level, but rather at the character level. The text is thus represented as vectors $[x_1,...,x_N]$ where $N$ is the number of characters in the sentence (the longest sentence has 45 characters) and $x_{1,.., N}$ are the one-hot encodings of length 27 (full English alphabet + space).  The total sequence length is thus different for each textual description - this is automatically handled by the toolkit using zero-padding and the creation of masks (see Section \ref{study}). A PCA visualization of \dataset Level 5 for both image and text modalities is shown in the Appendix in Fig. \ref{fig:pca}.

\subsection{Scalable complexity}
To enable finding the minimal functioning scenario for each model, the \dataset dataset can be generated in multiple levels of difficulty - by the difficulty we mean the number of semantic domains the model has to distinguish (rather than the size/dimensionality of the modalities). Altogether, there are 5 difficulty levels (see Fig. \ref{fig:data}):

\setlength{\parskip}{0pt} % 1ex plus 0.5ex minus 0.2ex}
\begin{itemize}[leftmargin=*]
\setlength{\itemsep}{-1pt}
    \item \textbf{Level 1} - the model only learns the shape names (e.g. \textit{square})
    \item \textbf{Level 2} - the model learns the shape and its size (e.g. \textit{small square})
    \item \textbf{Level 3} - the model has to learn the shape, its size and textured colour (\textit{small red square})
    \item \textbf{Level 4} - the model also has to learn the position (\textit{small red square at top left})
    \item \textbf{Level 5} - the model learns also the background shade (\textit{small red square at top left on dark})
\end{itemize}
\setlength{\parskip}{5pt} % 1ex plus 0.5ex minus 0.2ex}

%Since the above-mentioned levels are defined for the text modality, the user can also choose whether to keep the image complexity at the same level (e.g., for Level 1, only the shape varies with a fixed size, colour, position and background), or whether to use the full spectrum of attributes. The complexity of the bimodal dataset can be thus highly balanced (Level 5 for images and text) or highly imbalanced (Level 5 for images and Level 1 for text). 
We provide a detailed description of how to configure and generate the dataset on our Github repository \textsuperscript{\sref{git}}.\looseness=-1

\subsection{Automated evaluation}
\label{ssec:evl}
The rigid structure and synthetic origin of the \dataset dataset enable automatic evaluation of the coherence of the generated samples. The generated text can be tokenized and each word can be looked up in the dataset vocabulary based on its position in the sentence (for example, the first word always has to refer to the size in case of Level 5 of the dataset). The visual attributes of the generated images are estimated using pre-trained image classifiers specified for the given feature and difficulty level. Based on these metrics, it is possible to calculate the joint- and cross-generation accuracies for the given data pair - the image and text samples are considered coherent only if they correspond in all evaluated attributes. The fraction of semantically coherent (correct) generated pairs can thus serve as a percentual accuracy of the model. 

We report the joint- and cross-generation accuracies on three levels: \textit{Strict}, \textit{Features} and \textit{Letters} (only applicable for text outputs). The \textit{Strict} metrics measure the percentage of completely correct samples, i.e. there is zero error tolerance (all letters and all features in the image must be correct). The \textit{Features} metrics measure the ratio of correct features per sample (words for the text modality and visual features for the image modality), i.e. accuracy ratio 1.75(0.5)/5 on the level 5 means that on average 1.75 $\pm$ 0.5 features out of 5 are recognized correctly for each sample, and \textit{Letters} measures the average percentage of correct letters in the text outputs (see the Appendix Sec. \ref{ssec:eval} for a detailed description of the metrics).

% \begin{table}
%   \caption{Comparison of the currently used bimodal benchmark datasets for multimodal VAE evaluation. We compare whether both modalities are complex (e.g. the second modality is not a label), whether the modalities are from different domains (e.g. not two images), whether their empirical evaluation is unambiguous and whether they are scalable for various difficulty levels.}
%   \label{tab:datasettable}
%   \centering
%  \begin{tabularx}{\textwidth}{@{}{4cm}YYYYY@{}}
%     \toprule
%     Dataset     & Two Complex Modalities & Different Data Domains & Unambiguous Qualitative Evaluation &   Multiple Difficulty Levels  \\
%     \midrule
%     MNIST \citep{deng2012mnist} & $\times$ & N/A & \checkmark & $\times$   \\
%     FashionMNIST \citep{xiao2017fashion} &  $\times$ & N/A & \checkmark & $\times$ \\
%     MNIST-SVHN \citep{shi2019variational} & \checkmark & $\times$ & \checkmark &  $\times$     \\
%     CelebA \citep{liu2015faceattributes}  & \checkmark & \checkmark &  $\times$ &  $\times$  \\
%     CUB \citep{wah2011caltech} & \checkmark & \checkmark & $\times$ & $\times$ \\
%     \textbf{\dataset (ours)} & \textbf{\checkmark} & \checkmark & \textbf{\checkmark} & \textbf{\checkmark} \\
%     \bottomrule
%   \end{tabularx}
%  \vspace{-4mm} 
% \end{table}

\begin{table}
  \caption{\mycomment{Comparison of the currently used bimodal benchmark datasets for multimodal VAE evaluation. We compare the overall number of categories (we show the numbers for each level for CdSprites+), the dataset's overall size (we show the size of MNIST and the size of SVHN for MNIST-SVHN and the 5 Levels for CdSprites+), the domains of the two modalities, how the quality of the output can be evaluated (coherency is a more precise estimate compared to the vague canonical correlation analysis (CCA)) and the number of difficulty levels.}}
  \label{tab:datasettable}
  \centering
   \fontsize{10}{10}\selectfont % Adjust the font size here
\begin{tabular}{>{\centering\arraybackslash}p{2.7cm}>{\centering\arraybackslash}p{2cm}>{\centering\arraybackslash}p{1.8cm}>{\centering\arraybackslash}p{1.7cm}>{\centering\arraybackslash}p{1.7cm}>{\centering\arraybackslash}p{1cm}} 
    \toprule
    Dataset     & Categories & Size & Data Domains & Qualitative Evaluation &   Difficulty Levels  \\
    \midrule
    MNIST & 10 digits & 60k & image/label & coherency & 1   \\
    FashionMNIST & 10 articles  & 70k & image/label & coherency & 1 \\
    MNIST-SVHN & 10 digits  & 60k/600k & image/image & coherency &  1     \\
    CelebA  & 40 binary  & 200k & image/labels &  coherency & 1 \\
    CUB & 200 bird types  & 12k & image/text & CCA & 1 \\
    \textbf{\dataset (ours)} & 3/6/30/120/240  & 75/120/300/ 600/960k & image/text & coherency & 5 \\
        \bottomrule
  \end{tabular}
 \vspace{-4mm} 
\end{table}

\section{Benchmarking Toolkit}
\label{toolkit}

Our proposed toolkit \sref{git} was developed to facilitate and unify the evaluation and comparison of multimodal VAEs. Due to its modular structure, the tool enables adding new models, datasets, encoder and decoder networks or objectives without the need to modify any of the remaining functionalities. It is also possible to train unimodal VAEs to see whether the multimodal integration distorts the quality of the generated samples. 

The toolkit is written in Python, the models are defined and trained using the PyTorch Lightning library \citep{falcon2019pytorch}. For clarity, we directly state in the code which of the functions (e.g. various objectives) were taken from previous implementations so that the user can easily look up the corresponding mathematical expressions. 

Currently, the toolkit incorporates the MVAE \citep{wu2018multimodal}, MMVAE \citep{shi2019variational}, MoPoE-VAE \citep{sutter2021generalized} and DMVAE \citep{lee2021private} models. \added{All of the selected models are trained end-to-end and are able to handle missing modalities on the input, which we consider the basic requirement. You can see an overview of the main differences among the models adopted from \citet{suzuki2022survey} in the Appendix Table \ref{tab:modelcompar}.} 

The datasets supported by default are MNIST-SVHN, CUB, CelebA, Sprites (a trimodal dataset with animated game characters), FashionMNIST, PolyMNIST and our \dataset dataset. We also provide instructions on how to easily train the models on any new dataset. As for the encoder and decoder neural networks, we offer fine-tuned convolutional networks for image data (for all of the supported datasets) and several Transformer networks aimed at sequential data such as text, image sequences or actions (these can be robotic, human or any other kind of actions). All these components are invariant to the selected model and changing them requires only a change in the config file.

Although we are continuously extending the toolkit with new models and functionalities, the main emphasis is placed on providing a tool that any scientist can easily adjust for their own experiments. The long-term effort behind the project is to make the findings on multimodal VAEs reproducible and replicable. We thus also have tutorials on how to add new models or datasets to the toolkit in our GitHub documentation \footnote{\url{https://gabinsane.github.io/multimodal-vae-comparison}\label{docum}}.

%In the following subsections, we highlight some additional features that our toolkit offers to the user. 

\subsection{Experiment configuration}

The training setup and hyperparameters for each experiment can be defined using a YAML config file. Here the user can define any number and combination of modalities (unimodal training is also possible), modality-specific encoders and decoders, desired multimodal integration method, reconstruction loss term, objective (this includes various subsampling strategies due to their reported impact on the model convergence \citep{wu2018multimodal}, \citep{daunhawer2021limitations}) and several training hyperparameters (batch size, latent vector size, learning rate, seed, optimizer and more). For an example of the training config file, please refer to the toolkit repository \textsuperscript{\ref{git}}. 

\subsection{Evaluation methods}

We provide both dataset-independent and dataset-dependent evaluation metrics. The dataset-independent evaluation methods are the estimation of the test log-likelihood, plots with the KL divergence and reconstruction losses for each modality, and visualizations of the latent space using T-SNE and latent space traversals (reconstructions of latent vectors randomly sampled across each dimension). Examples are shown in the Appendix.

The dataset-dependent methods focus on the evaluation of the 4 criteria specified by \cite{shi2019variational}: \textit{coherent joint generation}, \textit{coherent cross-generation}, \textit{latent factorisation} and \textit{synergy}. For qualitative evaluation, joint- and cross-generated samples are visualised during and after training. For quantitative analysis of the generated images, we provide the Fréchet Inception Distance (FID) estimation. For our \dataset dataset, we offer an automated qualitative analysis of the joint and cross-modal coherence - the simplicity and rigid structure of the dataset enable binary (correct/incorrect) evaluation for each generated image (with the use of pre-trained classifiers) and also for each letter/word in the generated text. For more details, see Section \ref{ssec:evl}.

\begin{table}[!htp]
  \caption{\mycomment{Marginal log-likelihoods (lower is better) for the four models trained on the CelebA dataset using our toolkit. Variance calculated over 3 seeds is shown in brackets.}}
  \label{tab:celebares}
  \centering
 \begin{tabularx}{\textwidth}{@{}p{2.2cm}p{2.5cm}p{2.5cm}p{2.5cm}p{2.5cm}@{}}
    \toprule
     Metric & MMVAE & MVAE & MoPoE & DMVAE  \\
    \midrule
      $log{p(x_1)}$ & -6239.4 (1.2) &  \textbf{-6238.2 (1.6)} & -6241.3 (1.8) &  -6243.4 (1.3)   \\
     $log{p(x_1, x_2)}$ & \textbf{-6236.3 (0.9)} &  -6238.8 (1.1) & -6242.4 (2.3)& -6239.7 (1.2) \\
     $log{p(x_1|x_2)}$ & -6236.6 (1.6) &  -6235.7 (1.3) & \textbf{-6235.5 (1.2)} &  -6236.8 (1.4)  \\
    \bottomrule
  \end{tabularx}
 \vspace{-4mm} 
\end{table}

\subsection{Datasets for evaluation}

The users have the following options to evaluate their models:
\setlength{\parskip}{0pt} 
\begin{itemize}[leftmargin=*]
    \item \textbf{Using the implemented benchmark datasets.} By default, we provide 6 of the commonly used multimodal benchmark datasets such as MNIST-SVHN, CelebA, CUB etc. Training and testing on these models can be defined in a config.
    \item \textbf{Using the \dataset dataset.} We provide links for downloading all the 5 levels of our dataset. However, it is also possible to generate the data manually. \dataset dataset and its variations can be generated using a script provided in our toolkit - it allows the user to choose the complexity of the data (we provide 5 difficulty levels based on the number of included shape attributes), number of samples per category, specific colours and other features. The dataset generation requires only the standard computer vision libraries and can be generated in \mycomment{a matter of minutes or hours (based on the difficulty level)} on a CPU machine. You can read more about the dataset in Section \ref{datasetdesc}. 
    \item \textbf{Adding a new dataset.} Due to the modular structure of the toolkit, it is possible to add a new dataset class without disrupting other parts of the code. We provide documentation for how to do it in the GitHub pages linked in our repository \sref{git}.
\end{itemize}

\begin{table}[!htp]
\caption{\mycomment{Results for the four models trained on MNIST-SVHN using our toolkit. We show the digit classification accuracies (\%) of latent variables (\itshape{MNIST}) and (\itshape{SVHN}), and the probability of digit matching (\%) for cross- and joint-generation. The variance of the results calculated over 3 seeds is shown in brackets.}}
  \label{tab:mnistsvhnres}
  \centering
 \begin{tabularx}{\textwidth}{@{}p{2.8cm}p{1.6cm}p{1.6cm}p{2.2cm}p{2.2cm}p{1.9cm}@{}}
    \toprule
    Version & MNIST & SVHN & MNIST $\rightarrow$ SVHN & SVHN $\rightarrow$ MNIST & Joint \\
    \midrule
     MMVAE & 85.6 (5.2) & 86.4 (4.6) &  \textbf{85.7 (5.2)} & \textbf{84.5 (4.9)} & 45.3 (3.1) \\
     MVAE &  \textbf{91.5 (4.6)} & \textbf{88.2 (3.5)} & 95.4 (2.1) &  83.52 (3.6) & \textbf{55.6 (2.6)}  \\
     MoPoE & 86.6 (3.8) & 83.6 (4.3) & 81.2 (3.6) & 81.8 (2.7) & 43.2 (3.9) \\
     DMVAE & 78.27 (4.3) & 69.34 (5.6) &  84.5 (4.7) & 82.2 (3.1) & 44.9 (3.6)    \\
    \bottomrule
  \end{tabularx}
 \vspace{-4mm} 
\end{table}

\section{Benchmark Study}
\label{study}

In this section, we demonstrate the utility of the proposed toolkit and dataset on a set of experiments comparing two selected state-of-the-art multimodal VAEs. 

\subsection{Experiments}

We perform a set of experiments to compare the four widely discussed multimodal VAE approaches - MVAE \citep{wu2018multimodal}, with the Product-of-Experts multimodal integration, and MMVAE \citep{shi2019variational}, presenting the Mixture-of-Experts strategy, MoPoE \citep{sutter2021generalized} and DMVAE \citep{lee2021private}. Firstly, to verify that the implementations are correct, we replicated some experiments presented in the original papers inside our toolkit. We used the same encoder and decoder networks and training hyperparameters as in the original implementation and compared the final performance.\mycomment{The results are shown in Appendix Sec. \ref{sec:reproduce}. We then unified the implementations for all models so that they only differ in the modality mixing and trained them on the CelebA and MNIST-SVHN datasets. The results are shown in Tables \ref{tab:celebares} and \ref{tab:mnistsvhnres}}.

Next, we trained all four models on the \dataset dataset consecutively on all 5 levels of complexity and performed a hyperparameter grid search over the dimensionality of the latent space. You can find the used encoder and decoder architectures (fixed for all models) as well as the specific training details in Appendix Sec. \ref{ssec:confi}. We show the qualitative and quantitative results and discuss them in Section \ref{sec:results}.\looseness=-1

\begin{table}
  \caption{Results for the four models trained on all 5 levels of our \dataset dataset. \textit{Strict} refers to the percentage of completely correct samples (sample pairs in joint generation), \textit{Features} shows the ratio of correct features (i.e., 1.2 (0.1)/3 for Level 3 means that on average 1.2~$\pm$~0.1 features out of 3 are recognized correctly for each sample) and \textit{Letters} shows the mean percentage of correctly reconstructed letters (computed sample-wise). Standard deviations over 3 seeds are in brackets. For each model, we chose the most optimal latent space dimensionality \textit{Dims} (for DMVAE a sum of private and shared latents). Please see the Appendix for a detailed explanation of our metrics.}
  \label{tab:comparison}
  \footnotesize
  \centering
 \begin{tabularx}{\textwidth}{p{0.5cm}p{1cm}p{0.5cm}p{1.1cm}p{1.2cm}p{1.1cm}p{1.2cm}p{1.1cm}p{1cm}p{1.1cm}}%{@{}{0.5cm}YYYYYYYYY@{}}
    \toprule
    Level & Model  & Dims  & Txt$\rightarrow$Img Strict [\%] &  Txt$\rightarrow$Img Features   [ratio] &  Img$\rightarrow$Txt Strict [\%] & Img$\rightarrow$Txt Features [ratio] &  Img$\rightarrow$Txt Letters [\%] & Joint Strict [\%] & Joint Features [ratio]\\
 \toprule
   \multirow{2}{*}{1} & MMVAE &16 & 47 (14) & 0.5~(0.1)/1 & 64 (3) & \textbf{0.6~(0.0)/1} & \textbf{88 (2)} & \textbf{17 (10)} & \textbf{0.2~(0.1)/1} \\
                      & MVAE  &16 &  \textbf{52 (3)} & \textbf{0.5~(0.0)/1} & \textbf{63 (8)} & 0.6~(0.1)/1 & 86 (2) & 5 (9) & 0.1~(0.1)/1 \\
                      & MoPoE  &16 &  33 (3) & 0.3~(0.0)/1 & 10 (17) & 0.1~(0.2)/1 & 26 (7) & 16 (27) & 0.2~(0.3)/1 \\
                      & DMVAE  &30 & 33 (4) & 0.3~(0.0)/1 & 4 (5) & 0.0~(0.0)/1 & 25 (2) & 4 (6) & 0.0~(0.1)/1 \\
    \toprule
  \multirow{2}{*}{2} &MMVAE& 16&  \textbf{18 (4)} & 0.8~(0.1)/2 & 41 (20) & 1.4~(0.2)/2 & 85 (4) & \textbf{3 (3)} & \textbf{0.6~(0.1)/2} \\
                     &MVAE&12  & 16 (1) & 0.8~(0.0)/2 & \textbf{55 (27)} & \textbf{1.5~(0.3)/2} & \textbf{91 (6)} & 1 (1) & 0.3~(0.3)/2 \\
                     & MoPoE  &16 &  10 (3) & 0.8~(0.0)/2 & 8 (7) & 0.7~(0.1)/2 & 40 (4) & 1 (1) & 0.2~(0.1)/2 \\
                     & DMVAE  &30 &  15 (2) & 0.8~(0.0)/2 & 4 (1) & 0.4~(0.0)/2 & 30 (2) & 0 (0) & 0.2~(0.1)/2 \\

    \toprule
  \multirow{2}{*}{3} &MMVAE&16 & 6 (2) & 1.2~(0.2)/3 & 2 (3) & 0.6~(0.2)/3 & 31 (5) & 0 (0) & 0.4~(0.1)/3 \\
                     &MVAE&32  &\textbf{8 (2)} & \textbf{1.3~(0.0)/3} & \textbf{59 (4)} & \textbf{2.5~(0.1)/3} & \textbf{93 (1)} & 0 (0) & \textbf{0.5~(0.1)/3} \\
                     & MoPoE  &24 &  7 (4) & 1.3~(0.1)/3 & 0 (0) & 0.7~(0.1)/3 & 32 (0) & 0 (0) & 1.1~(0.1)/3 \\
                     & DMVAE  &30 & 4 (0) & 1.2~(0.0)/3 & 0 (0) & 0.4~(0.1)/3 & 22 (2) & \textbf{1 (1)} & \textbf{0.5~(0.1)/3} \\
    \toprule
  \multirow{2}{*}{4} &MMVAE&24 &  \textbf{3 (3)} & \textbf{1.7~(0.4)/4} & \textbf{1 (2)} & 0.7~(0.4)/4 & \textbf{27 (9)} & 0 (0) & 0.5~(0.2)/4 \\
                     &MVAE&16   &  0 (0) & 1.3~(0.0)/4 & 0 (0) & 0.2~(0.3)/4 & 16 (5) & 0 (0) & 0.3~(0.6)/4    \\
                     & MoPoE  &24 &  2 (1) & 1.4~(0.0)/4 & 0 (0) & \textbf{0.7~(0.1)/4} & 21 (3) & 0 (0) & 0.1~(0.2)/4 \\
                     & DMVAE  &30 &  1 (1) & 1.4~(0.0)/4 & 0 (0) & 0.5~(0.1)/4 & 18 (1) & 0 (0) & \textbf{0.5~(0.1)/4} \\
      \toprule
  \multirow{2}{*}{5} &MMVAE&24&  0 (0) & 1.8~(0.0)/5 & 0 (0) & 0.1~(0.1)/5 & 13 (2) & 0 (0) & 0.4~(0.1)/5  \\
                    &MVAE &16 & 0 (0) & 1.8~(0.0)/5 & 0 (0) & 0.6~(0.0)/5 & \textbf{27 (1)} & 0 (0) & 0.2~(0.2)/5 \\
                    & MoPoE  &24 & 0 (0) & 1.8~(0.0)/5 & 0 (0) & \textbf{0.7~(0.0)/5} & 17 (1) & 0 (0) & \textbf{1.0~(0.0)/5} \\
                    & DMVAE  &30& 0 (0) & 1.8~(0.0)/5 & 0 (0) & 0.6~(0.1)/5 & 18 (2) & 0 (0) & 0.7~(0.1)/5 \\

    \bottomrule
  \end{tabularx}
  \vspace{-6mm}
\end{table}

\begin{figure}[!htp]
\centering
\setlength{\belowcaptionskip}{-16pt}
\includegraphics[width=.94\textwidth]{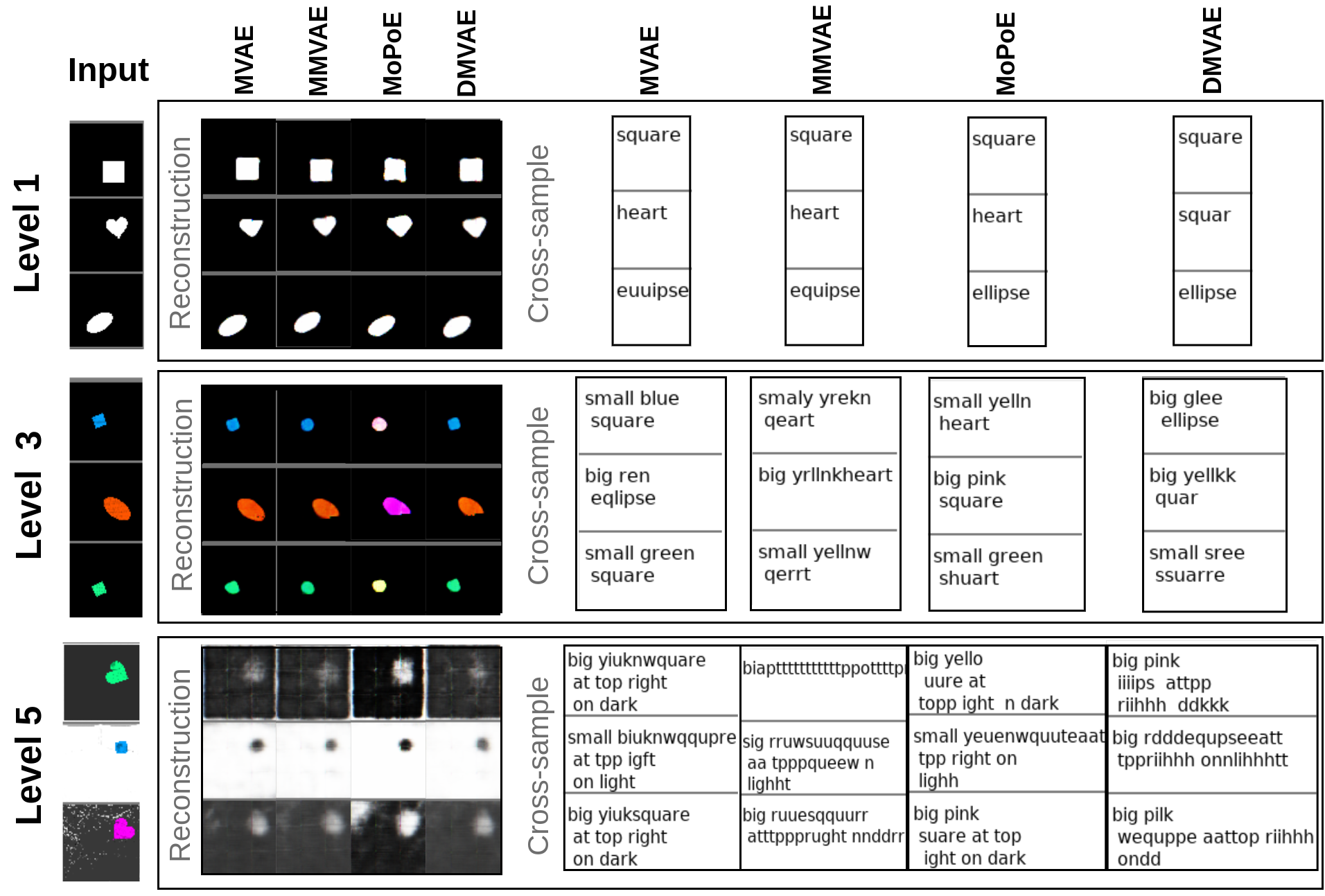}
\caption{Qualitative results of the MVAE, MMVAE, MoPoE and DMVAE models trained on Level 1, 3 and~5 of our \dataset dataset. We show first the reconstructions of the input image, then the captions obtained by cross-sampling. }
\label{fig:datasetres}
\end{figure}

% while Wu and Goodman \citep{wu2018multimodal} train the image reconstruction using binary cross-entropy loss (BCE):

% \begin{equation}
% \label{eqn:bce}
% \resizebox{0.5\hsize}{!}{L_{BCE}=-\frac{1}{M}\sum_{i=1}^{M}[\boldsymbol{y_i}log\boldsymbol{\hat{y}_i} + (1-\boldsymbol{y_i})log(1-\boldsymbol{\hat{y}_i})]},
% \end{equation}

% where $\hat{y}_i$ is the prediction for each of the \textit{M} RGB values of the image and $y_i$ is the ground truth normalized between 0 and 1. 

% We report and discuss the results in the following section. 

\subsection{Results}
\label{sec:results}

The detailed results for all experiments (including other datasets such as Sprites, MNIST-SVHN etc.) can be found in Appendix Sec. \ref{sec:resultsdetailed}. Here we demonstrate the utility of our toolkit and dataset by comparing the MVAE, MMVAE, MoPoE and DMVAE models on the \dataset dataset consecutively on 5 levels of difficulty. For an overview of what attributes each of the levels includes, please see Section \ref{datasetdesc}. In Fig.~\ref{fig:datasetres}, we show the qualitative results for levels 1, 3 and 5 of the dataset for the four models. We show both the reconstructed and cross-generated (conditioned on the other modality) samples for images and text. Moreover, we report the cross- ($Img \rightarrow Txt$ and $Txt  \rightarrow Img$) and joint- ($Joint$) coherency accuracies for all levels in Table \ref{tab:comparison}. The \textit{Strict} metrics show the percentage of completely correct samples, while \textit{Features} and \textit{Letters} show the average proportion of correct words (or visual features for image) or letters per sample. 

Based on Table \ref{tab:comparison}, MVAE and MMVAE outperform MoPoE and DMVAE in almost all categories for Levels 1 and 2. While the difference between MVAE and MMVAE is not large in Level 1, MVAE produces far more precise and stable text reconstructions up until Level 3. This would be in accordance with the results by \cite{kutuzova2021multimodal}, who showed that the PoE approach outperforms the MoE approach on datasets with a multiplicative ("AND") combination of modalities, which is also the case of our \dataset dataset.  
The most prominent trend that can be seen both in Table \ref{tab:comparison} and Fig.~\ref{fig:datasetres} is the gradual performance decline across individual Levels. This is the expected and desirable outcome since only a challenging benchmark allows us to distinguish the difference among the evaluated models. For more details on the results, please see Appendix Sec. \ref{sec:resultsdetailed}.

 \section{Conclusions}

In this work, we present a benchmarking toolkit and a CdSprites+ ({\it{Captioned disentangled Sprites+}}) dataset for a systematic evaluation and comparison of multimodal variational autoencoders. The tool enables the user to easily configure the experimental setup by specifying the dataset, encoder and decoder architectures, multimodal integration strategy and the desired training hyperparameters all in one config. The framework can be easily extended for new models, datasets, loss functions or the encoder and decoder architectures without the need to restructure the whole environment. In its current form, it includes 4 state-of-the-art models and 6 commonly used datasets.

Furthermore, the proposed synthetic bimodal dataset offers an automated evaluation of the cross-generation and joint-generation capabilities of the multimodal VAE models on 5 different scales of complexity. We also offer several automatic visualization modules that further inform on the latent factorisation of the trained models. We have evaluated the incorporated models on our CdSprites+ dataset and the results are publicly displayed in a "leaderboard" table on the GitHub repository, which will be gradually extended when new models and results are available \textsuperscript{\ref{git}}. 

The limitation of our \dataset dataset is its synthetic nature and consequent simplicity compared to real-world data. This is a trade-off for the constrained variability allowing a reliable and fully automated evaluation. Please note that our dataset is aimed as a diagnostic tool allowing comparison of performance nuances across various models. We still encourage the researchers to train their models on real-world datasets (which we also include in our toolkit) at the same time to show their full capacity. 

%We wish to encourage the users to build on and extend our framework so that it can serve as a benchmarking tool for reproducible and replicable research which is currently highly needed in the area of multimodal VAEs.

\bibliography{iclr2024_conference}
\bibliographystyle{iclr2024_conference}

\appendix
\section{Appendix}

 In Section \ref{sec:cdsprites}, we provide additional details for the \dataset dataset. In Section \ref{sec:resultsdetailed}, we describe the technical information and detailed results for the experiments presented in the paper \mycomment{and Section \ref{sec:reproduce} reports the reproduced results from original papers using our toolkit}. 

\subsection{\dataset dataset statistics}
\label{sec:cdsprites}

The presented version of the benchmark dataset comprises 5 different levels of difficulty, where each level varies in the number of included features (see Table \ref{tab:stats} for their overview). The default size of the dataset is depicted in Table \ref{tab:stats} for each level, although the user can easily generate a larger set of the data. The scripts for calculation of the cross- and joint- coherency accuracies use separate batches of testing data provided in the toolkit.

In all levels, the source of noise in the images is their random position and rotation - in levels 1-4, the shapes are located around the whole image with a variance of 25 pixels along both $x$ and $y$ axes. In Level 5, the shapes are shifted in random quadrants where their position also varies with a variance of 8 pixels. The positions are configured so that the whole shapes are always fitting the image. In Levels 2-5 where we vary the size, the proportion of the small objects to the big objects is 1:5.

\begin{figure}[!htp]
\centering
\includegraphics[width=1\textwidth]{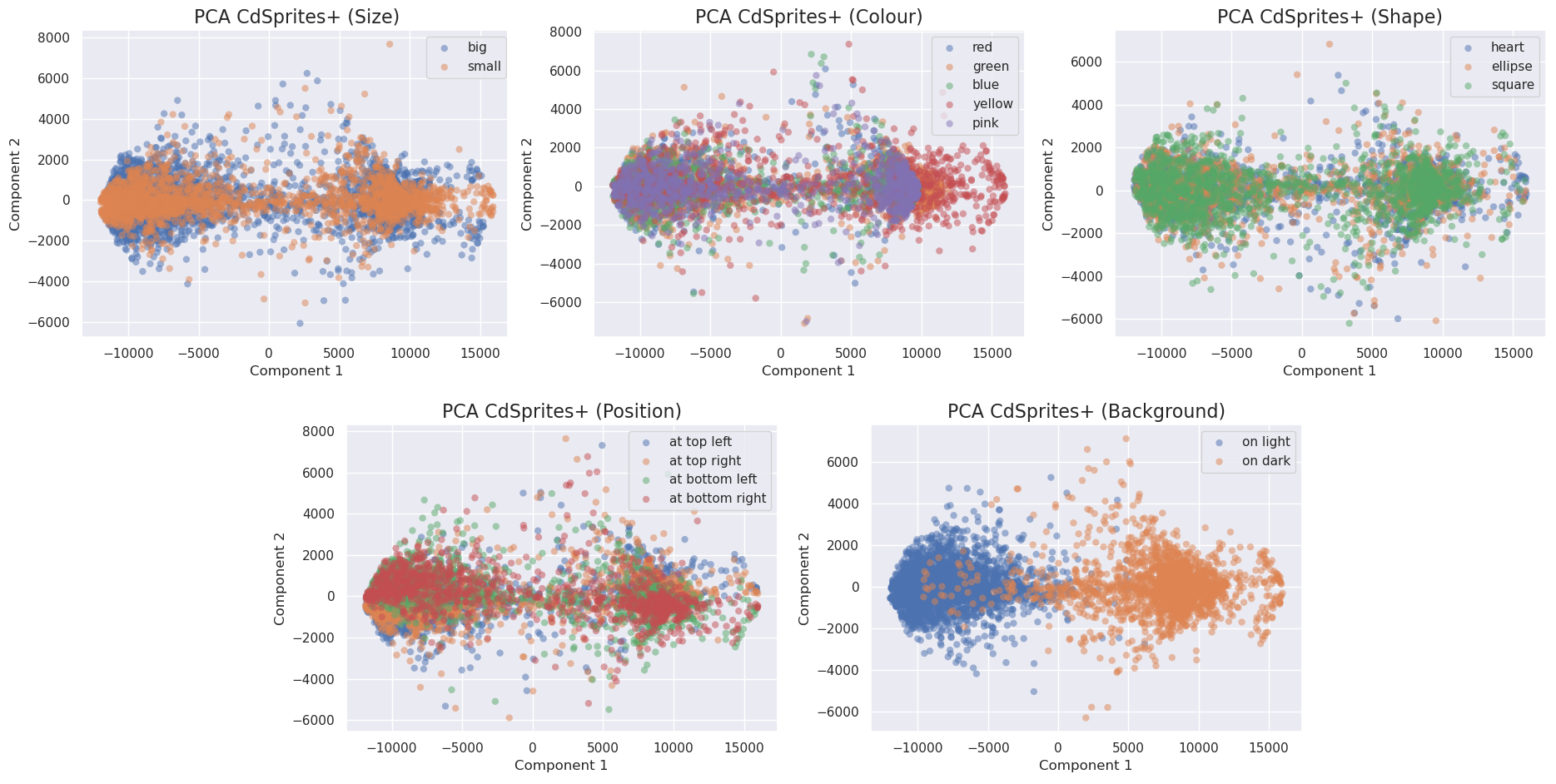}
\caption{PCA calculated on the images in our \dataset dataset, Level 5. We show a separate figure for each of the 5 features (size, shape, position and colour).}
\label{fig:pca}
\end{figure}

You can also see a PCA visualization of the \dataset Level 5 dataset in Fig. \ref{fig:pca}.

\begin{table}[htp!]
  \caption{Statistics of the \dataset benchmark dataset. We show the number of train/validation samples and the number of various shapes, colours, object poses (meaning quadrants which are distinguished in captions) and backgrounds used in each difficulty level. The text captions only describe features that vary (e.g. in level 1, the text descriptions only include the shape name). The colours and backgrounds are all textured when they vary.}

  \label{tab:stats}
  \small
  \centering
 \begin{tabularx}{\textwidth}{@{}p{0.8cm}p{1.8cm}p{2.4cm}p{1.0cm}p{1.0cm}p{1.0cm}p{1.0cm}p{1.5cm}@{}}
    \toprule
    Level & Train Samples & Validation Samples & Shapes & Sizes & Colours & Positions & Backgrounds \\
 \toprule
    1 & 67 500 & 7 500 & 3 & 1 & 1 & 1 & 1 \\
    2 & 108 000 & 12 000 & 3 & 2 & 1 & 1 & 1 \\
    3 & 270 000 & 30 000 & 3 & 2 & 5 & 1 & 1 \\
    4 & 540 000 & 60 000 & 3 & 2 & 5 & 4 & 1 \\
    5 & 864 000 & 96 000 & 3 & 2 & 5 & 4 & 2 \\
      \end{tabularx}
\end{table}

\subsubsection{\mycomment{Using character-wise embeddings}}
    \mycomment{We choose to use character-wise embeddings for CdSprites+ rather than word embeddings. While this choice was made to increase the difficulty of the text modality in our dataset, character embeddings have been recently used also in several other works as this approach brings specific advantages.}

\mycomment{Firstly, character-wise embedding does not require a pre-defined vocabulary of possible input words. This can be useful e.g. in incremental learning scenarios where the whole vocabulary is not known prior to training beginning.} \mycomment{Secondly, the model can be tested for robustness after training by inputting sentences with misspelt words (e.g., “sqaare” instead of “square”) to see if the model can generate correct images. With word-level embedding, this is not possible as replacing entire words will change the feature or create a nonsensical query (e.g., “left square” instead of “blue square”).}

\mycomment{Please note that we expect the users to use the same encoder and decoder networks (i.e. character transformer networks) for the CdSprites+ benchmark to provide a restricted and fair comparison to other models. Should the users want to use CdSprites+ outside our toolkit for their custom evaluation, they can as well use word-level embeddings as we provide raw strings for the CdSprites+ text modality.}

\subsection{Benchmark study results}
Here we provide the specific training configuration and hyperparameters used for the experiments on the \dataset dataset as listed in the paper. We also report the detailed results for hyperparameter grid search in terms of the cross- and joint-generation accuracies. 

\subsubsection{Training configuration}
\label{ssec:confi}
All our experiments were trained with the GeForce GTX 1080 and NVIDIA Tesla V100 GPU cards, the mean computation \mycomment{times for training and inference are shown in Table \ref{tab:times}}. We used the Adam optimizer, the learning rate of $1e^{-4}$ and all experiments were repeated for 5 seeds (we report standard deviations for the results in the tables). We trained for 150 epochs for Levels 1 and 2 and for 250 epochs in the case of Levels 3-5. In the hyperparameter grid search, we varied the latent dimensionality (16, 24, 32) for all 5 dataset levels and the MVAE, MMVAE and MoPoE models. In the case of DMVAE, the latent dimensionality was different as there are private (modality-dependent) and shared latents. We thus chose different values for the comparison. We used a fixed value of 10 for both private latents and varied the shared latents with values 10, 16 and 24. In Tables 1-5, we show this as the total number of latent dimensions, i.e. 30 (10 shared and $2\times 10$ private), 36 (16 shared and $2\times 10$ private) and 46 (24 shared and $2\times 10$ private).  

We used the default training dataset size and validation split as reported in the statistics Table \ref{tab:stats}. In Tables \ref{tab:level1}, \ref{tab:level2}, \ref{tab:level3}, \ref{tab:level4} and \ref{tab:level5}, we show results for the MVAE, MMVAE, DMVAE and MoPoE models and the compared latent dimensionalities. Standard deviations over 5 seeds are shown in brackets.

\begin{table}
  \caption{Training and inference times for each model trained on our \dataset dataset. The models were trained for 150 epochs on Levels 1-2 and for 250 epochs on Levels 3-5, we thus show these times separately. We show the mean values over all seeds and different latent dimensionalities, the standard deviation is shown as $\pm$.}
    \label{tab:times}
  \small
  \centering
 \begin{tabularx}{\textwidth}{@{}p{2cm}p{2cm}p{2.8cm}p{2.8cm}p{2.3cm}@{}}
    \toprule
    Model & Per epoch (s) & Per training (min) \newline  (Levels 1-2)  & Per training (min) \newline (Levels 3-5) & Inference time (s)  \\
 \toprule
    MMVAE & 203  $\pm$ 2 & 525  $\pm$ 8 & 860  $\pm$ 10 & 152  $\pm$ 6   \\
    MVAE & 150  $\pm$ 20 & 397 $\pm$ 25 & 645 $\pm$ 32 & 135 $\pm$ 19 \\
    MoPoE & 127 $\pm$ 12 & 324 $\pm$ 9 & 542 $\pm$ 10 & 126 $\pm$ 11  \\
    DMVAE & 200 $\pm$ 3 & 500 $\pm$ 6 & 841 $\pm$ 7 & 148 $\pm$ 8\\
      \end{tabularx}
\end{table}

\subsubsection{Used architecture}

For all evaluated models, we used the standard ELBO loss function with the $\beta$ parameter fixed to 1. For the MVAE [5] model, we used the sub-sampling approach where the model is trained on all subsets of modalities (i.e. images only, text only and images+text). For the image encoder and decoder, we used 4 fully connected layers with ReLU activations. In the case of the text, we used a Transformer network with 8 layers, 2 attention heads, 1024 hidden features and a dropout of 0.1. 

\subsubsection{Evaluation metrics}
\label{ssec:eval}

After training, we used the script for automated evaluation (provided in our toolkit) to compute the cross- and joint-coherency of the models. For cross-coherency, we generated a \mycomment{10000}-sample test dataset using the dataset generator and used first the images, and then captions as input to the model to reconstruct the missing modality. For joint coherency, we generated 1000 traversal samples over each dimension of the latent space (i.e. 32000 samples for a 32-D latent space) and fed these latent vectors into the models to reconstruct both captions and images.

For both the cross- and joint-coherencies, we report the following metrics: \textit{Strict}, \textit{Feat}, and \textit{Letters} to provide more information on what the models are capable to do. In the first ( \textit{Strict}) metrics, we considered the text sample as accurate only if all letters in the description were 100~\% accurate, i.e. we did not tolerate any noise. For the image outputs, we considered the images as correct only if all the attributes for the given difficulty level could be detected using our pre-trained classifiers (i.e. correct classification for the shape, colour, size, position or background). For joint coherency, we considered the generated pair as correct only when both the image and captions fulfilled these criteria and were semantically matching. 

For the feature-level metrics, we calculated the percentage of correctly reconstructed/generated features (e.g. whole words or image attributes such as shape) and reported the mean percentage of correct features per sample. For the image-caption cross-generation accuracy, we also calculated the average percentage of correct letters per output sample.

In the following section, we report the mean accuracies for both cross- and joint-coherency - these numbers describe the proportion of the correct outputs to all outputs.

\begin{table}
  \caption{Level 1 comparison of accuracies for the four evaluated models trained on our \dataset dataset. \textit{Strict} refers to percentage of completely correct samples (sample pairs in joint generation), \textit{Feats} shows the average percentage of correct features (Level 1 has only 1 feature and \textit{Feats} and \textit{Strict} are thus the same) and \textit{Letters} shows the mean percentage of correctly reconstructed letters.(\textit{Dim}) is the latent space dimensionality.}
  \label{tab:level1}
  \small

\begin{tabularx}{\textwidth}{@{}lYYYYYYY@{}}
    \toprule
    Model (Dim)  & Txt$\rightarrow$Img Strict \% &  Txt$\rightarrow$Img Feats \% &  Img$\rightarrow$Txt Strict \% & Img$\rightarrow$Txt Feats \% &  Img$\rightarrow$Txt Letters \%& Joint Strict \% & Joint Feats \%\\
 \toprule
   MMVAE (16-D) & 47 (14) & N/A  & \textbf{64 (3)} & N/A  & \textbf{88 (2)} & \textbf{17 (10)} & N/A  \\
   MVAE (16-D) &    \textbf{52 (3)} & N/A  & 63 (8) & N/A  & 86 (2) & 5 (9) & N/A \\
   DMVAE (30-D) & 33 (4) & N/A & 4 (5) & N/A & 25 (2) & 4 (6) & N/A  \\
   MoPoE (16-D) & 33 (3) & N/A  & 10 (17) &N/A  & 26 (7) & 16 (27) & N/A  \\
 \toprule
   MMVAE (24-D) &55 (15) & N/A & 42 (3) & N/A & 31 (12) & 0 (0) & N/A    \\
   MVAE (24-D) & \textbf{55 (4)} &  N/A & \textbf{61 (3)} & N/A & \textbf{82 (1)} & 3 (2) &  N/A   \\
   DMVAE (36-D) & 36 (1) &  N/A & 3 (3) &  N/A & 21 (2) & \textbf{9 (13)} &  N/A  \\
   MoPoE (24-D) & 35 (3) & N/A  & 4 (2) & N/A  & 24 (6) & 1 (1) & N/A   \\
 \toprule
   MMVAE (32-D) & 48 (3) & N/A & 36 (2) & N/A & 26 (2) & 0 (0) & N/A  \\
   MVAE (32-D) &  \textbf{53 (5)} & N/A & \textbf{60 (2)} & N/A & \textbf{82 (2)} & \textbf{1 (1)} & N/A  
 \\
    DMVAE (46-D) & 34 (2) & N/A & 3 (2) & N/A & 20 (9) & 0 (0) & N/A  \\
   MoPoE (32-D) & 36 (5) & N/A & 2 (1) & N/A & 23 (7) & 0 (0) & N/A  \\
    \bottomrule
  \end{tabularx}
\end{table}

\begin{table}
  \caption{Level 2 comparison of accuracies for the 4 models trained on our \dataset dataset. \textit{Strict} refers to the percentage of completely correct samples (sample pairs in joint generation), \textit{Feats} shows the average percentage of correct features (Level 2 has 2 features) and \textit{Letters} shows the mean percentage of correctly reconstructed letters.(\textit{Dim}) is the latent space dimensionality.}
  \label{tab:level2}
  \small
  \centering
\begin{tabularx}{\textwidth}{@{}lYYYYYYY@{}}
    \toprule
    Model (Dim)  & Txt$\rightarrow$Img Strict \% &  Txt$\rightarrow$Img Feats \% &  Img$\rightarrow$Txt Strict \% & Img$\rightarrow$Txt Feats \% &  Img$\rightarrow$Txt Letters \%& Joint Strict \% & Joint Feats \%\\
 \toprule
   MMVAE (16-D) & \textbf{18 (4)} & 0.8~(0.1)/2 & 41 (20) & 1.4~(0.2)/2 & 85 (4) & \textbf{3 (3)} & \textbf{0.6~(0.1)/2} \\
   MVAE (16-D) & 16 (1) & 0.8~(0.0)/2 & \textbf{55 (27)} & \textbf{1.5~(0.3)/2} & \textbf{91 (6)} & 1 (1) & 0.3~(0.3)/2    \\
   DMVAE (30-D) &  15 (2) & 0.8~(0.0)/2 & 4 (1) & 0.4~(0.0)/2 & 30 (2) & 0 (0) & 0.2~(0.1)/2 \\
   MoPoE (16-D) &  10 (3) & 0.8~(0.0)/2 & 8 (7) & 0.7~(0.1)/2 & 40 (4) & 1 (1) & 0.2~(0.1)/2   \\
    \toprule
   MMVAE (24-D) & 17 (5) & 0.4~(0.0)/2 & 16 (0) & 0.4~(0.0)/2 & 40 (2) & 1 (1) & 0.2~(0.0)/2 \\
       MVAE (24-D) & 16 (3) & 0.4~(0.0)/2 & \textbf{52 (9)} & \textbf{0.8~(0.0)/2} & \textbf{86 (1)} & \textbf{5 (6)} & 0.3~(0.0)/2  \\
      DMVAE (36-D) & \textbf{18 (2)} & \textbf{0.9~(0.0)/2} & 5 (1) & 0.4~(0.0)/2 & 24 (1) & 0 (0) & 0.2~(0.2)/2 \\
   MoPoE (24-D) & 8 (3) & 0.8~(0.0)/2 & 13 (3) & 0.8~(0.1)/2 & 35 (3) & 1 (1) & \textbf{0.5~(0.1)/2}  \\
    \toprule
   MMVAE (32-D) & \textbf{17 (1)} & 0.4~(0.0)/2 & 16 (0) & 0.5~(0.0)/2 & 43 (2) & 0 (0) & 0.1~(0.0)/2  \\
   MVAE (32-D) & 16 (4) & 0.8~(0.1)/2 & \textbf{40 (13)} & \textbf{1.8~(0.1)/2} & \textbf{87 (1)} & \textbf{11 (9)} & \textbf{0.8~(0.0)/2}\\
      DMVAE (46-D) & \textbf{17 (1)} & \textbf{0.8~(0.0)/2} & 3 (1) & 0.4~(0.1)/2 & 24 (2) & 0 (0) & 0.1~(0.1)/2 \\
   MoPoE (32-D) &7 (2) & \textbf{0.8~(0.0)/2} & 10 (8) & 0.8~(0.1)/2 & 33 (1) & 0 (0) & 0.3~(0.2)/2   \\
    \bottomrule
  \end{tabularx}
%\end{table}
%\begin{table}
  \caption{Level 3 comparison of accuracies for the 4 models trained on our \dataset dataset. \textit{Strict} refers to the percentage of completely correct samples (sample pairs in joint generation), \textit{Feats} shows the average percentage of correct features (Level 3 has 3 features) and \textit{Letters} shows the mean percentage of correctly reconstructed letters.(\textit{Dim}) is the latent space dimensionality.}
  \label{tab:level3}
  \small
  \centering
\begin{tabularx}{\textwidth}{@{}lYYYYYYY@{}}
    \toprule
    Model (Dim)  & Txt$\rightarrow$Img Strict \% &  Txt$\rightarrow$Img Feats \% &  Img$\rightarrow$Txt Strict \% & Img$\rightarrow$Txt Feats \% &  Img$\rightarrow$Txt Letters \%& Joint Strict \% & Joint Feats \%\\
 \toprule
   MMVAE (16-D) & 6 (2) & 1.2~(0.2)/3 & 2 (3) & 0.6~(0.2)/3 & 31 (5) & 0 (0) & 0.4~(0.1)/3 \\
   MVAE (16-D) & 6 (0) & 1.3~(0.0)/3 & \textbf{22 (12)} & \textbf{2.1~(0.1)/3} & \textbf{85 (3)} & 0 (0) & \textbf{0.5~(0.1)/3}   \\
   DMVAE (30-D) &  4 (0) & 1.2~(0.0)/3 & 0 (0) & 0.4~(0.1)/3 & 22 (2) & \textbf{1 (1)} & \textbf{0.5~(0.1)/3} \\
   MoPoE (16-D) & \textbf{6 (1)} & \textbf{1.6~(0.0)/3} & 0 (0) & 0.1~(0.1)/3 & 21 (5) & 0 (0) & 0.0~(0.0)/3   \\
    \toprule
   MMVAE (24-D) & 4 (3) & 1.2~(0.3)/3 & 1 (1) & 0.8~(0.2)/3 & 31 (6) & 0 (0) & 0.3~(0.0)/3  \\
   MVAE (24-D) & \textbf{7 (1)} & \textbf{1.3~(0.0)/3} & \textbf{45 (3)} & \textbf{2.4~(0.0)/3} & \textbf{91 (1)} & 0 (0) & \textbf{0.6~(0.0)/3}  \\
   DMVAE (36-D) & 3 (2) & 1.1~(0.1)/3 & 0 (0) & 0.2~(0.0)/3 & 18 (1) & 0 (0) & 0.1~(0.0)/3 \\
   MoPoE (24-D) &  7 (4) & 1.3~(0.1)/3 & 0 (0) & 0.7~(0.1)/3 & 32 (0) & 0 (0) & 1.1~(0.1)/3   \\
    \toprule
   MMVAE (32-D) & 5 (3) & 1.1~(0.1)/3 & 1 (1) & 0.6~(0.1)/3 & 28 (2) & 0 (0) & 0.0~(0.0)/3   \\
   MVAE (32-D) &  \textbf{8 (2)} & 1.3~(0.0)/3 & \textbf{59 (4)} & \textbf{2.5~(0.1)/3} & \textbf{93 (1)} & 0 (0) & \textbf{0.5~(0.1)/3} \\
   DMVAE (46-D) & 5 (1) & 1.1~(0.0)/3 & 0 (0) & 0.1~(0.1)/3 & 15 (1) & 0 (0) & 0.1~(0.0)/3 \\
   MoPoE (32-D) & \textbf{8 (2)} & \textbf{1.5~(0.1)/3} & 0 (1) & 0.6~(0.1)/3 & 28 (1) & 0 (0) & 0.5~(0.2)/3   \\
   
    \bottomrule
  \end{tabularx}
%\end{table}
%\begin{table}
  \caption{Level 4 comparison of accuracies for the 4 models trained on our \dataset dataset. \textit{Strict} refers to the percentage of completely correct samples (sample pairs in joint generation), \textit{Feats} shows the average percentage of correct features (Level 4 has only 4 features) and \textit{Letters} shows the mean percentage of correctly reconstructed letters.(\textit{Dim}) is the latent space dimensionality.}
  \label{tab:level4}
  \small
  \centering
\begin{tabularx}{\textwidth}{@{}lYYYYYYY@{}}
    \toprule
    Model (Dim)  & Txt$\rightarrow$Img Strict \% &  Txt$\rightarrow$Img Feats \% &  Img$\rightarrow$Txt Strict \% & Img$\rightarrow$Txt Feats \% &  Img$\rightarrow$Txt Letters \%& Joint Strict \% & Joint Feats \%\\
 \toprule
   MMVAE (16-D) &2 (0) & \textbf{1.6~(0.2)/4} & 0 (0) & 0.4~(0.4)/4 & 15 (3) & 0 (0) & 0.1~(0.1)/4 \\
   MVAE (16-D) &   0 (0) & 1.3~(0.0)/4 & 0 (0) & 0.2~(0.3)/4 & 16 (5) & 0 (0) & 0.3~(0.6)/4   \\
   DMVAE (30-D) & 1 (1) & 1.4~(0.0)/4 & 0 (0) & \textbf{0.5~(0.1)/4} & \textbf{18 (1)} & 0 (0) & \textbf{0.5~(0.1)/4}  \\
   MoPoE (16-D) & \textbf{3 (1)} & \textbf{1.6~(0.2)/4} & 0 (0) & \textbf{0.5~(0.1)/4} & 16 (3) & 0 (0) & 0.1~(0.1)/4  \\
    \toprule
   MMVAE (24-D) & 3 (3) & \textbf{1.7~(0.4)/4} & \textbf{1 (2)} & 0.7~(0.4)/4 & \textbf{27 (9)} & 0 (0) & \textbf{0.5~(0.2)/4} \\
   MVAE (24-D) &  \textbf{4 (1)} & 1.2~(0.1)/4 & 0 (1) & \textbf{2.4~(0.0)/4} & 14 (1) & 0 (0) & 0.2~(0.1)/4  \\
   DMVAE (36-D) &0 (1) & 1.3~(0.0)/4 & 0 (0) & 0.2~(0.0)/4 & 14 (1) & 0 (0) & 0.2~(0.0)/4 \\
   MoPoE (24-D) &  2 (1) & 1.4~(0.0)/4 & 0 (0) & 0.7~(0.1)/4 & 21 (3) & 0 (0) & 0.1~(0.2)/4   \\
    \toprule
   MMVAE (32-D) & 1 (1) & 1.6~(0.0)/4 & 0 (0) & 0.9~(0.0)/4 & 21 (0) & 0 (0) & 0.2~(0.0)/4  \\
   MVAE (32-D) & 2 (1) & 1.1~(0.1)/4 & 0 (1) & \textbf{1.1~(0.0)/4} & 12 (3) & 0 (0) & \textbf{0.4~(0.2)/4} \\
   DMVAE (46-D) & 1 (1) & 1.2~(0.0)/4 & 0 (0) & 0.1~(0.0)/4 & 14 (1) & 0 (0) & 0.1~(0.0)/4 \\
   MoPoE (32-D) & \textbf{4 (0)} & \textbf{1.7~(0.1)/4} & 0 (0) & 0.5~(0.3)/4 & \textbf{20 (3)} & 0 (0) & 0.2~(0.2)/4   \\

    \bottomrule
  \end{tabularx}
\end{table}

\begin{table}
  \caption{Level 5 comparison of accuracies for the 4 models trained on our \dataset dataset. \textit{Strict} refers to the percentage of completely correct samples (sample pairs in joint generation), \textit{Feats} shows the average percentage of correct features (Level 5 has 5 features) and \textit{Letters} shows the mean percentage of correctly reconstructed letters. (\textit{Dim}) is the latent space dimensionality.}
  \label{tab:level5}
  \small
  \centering
\begin{tabularx}{\textwidth}{@{}lYYYYYYY@{}}
    \toprule
    Model (Dim)  & Txt$\rightarrow$Img Strict \% &  Txt$\rightarrow$Img Feats \% &  Img$\rightarrow$Txt Strict \% & Img$\rightarrow$Txt Feats \% &  Img$\rightarrow$Txt Letters \%& Joint Strict \% & Joint Feats \%\\
 \toprule
   MMVAE (16-D) &0 (0) & 1.8~(0.0)/5 & 0 (0) & 0.4~(0.2)/5 & 16 (0) & 0 (0) & 0.7~(0.4)/5 \\
   MVAE (16-D) & 0 (0) & 1.8~(0.0)/5 & 0 (0) & \textbf{0.6~(0.0)/5} & \textbf{27 (1)} & 0 (0) & 0.2~(0.2)/5  \\
   DMVAE (30-D) & 0 (0) & 1.8~(0.0)/5 & 0 (0) & 0.6~(0.1)/5 & 18 (2) & 0 (0) & \textbf{0.7~(0.1)/5} \\
   MoPoE (16-D) & 0 (0) & 1.8~(0.0)/5 & 0 (0) & 0.3~(0.2)/5 & 15 (1) & 0 (0) & 0.5~(0.7)/5  \\
    \toprule
   MMVAE (24-D) & 0 (0) & 1.8~(0.0)/5 & 0 (0) & 0.6~(0.1)/5 & 17 (2) & 0 (0) & 0.5~(0.1)/5  \\
   MVAE (24-D) &  0 (0) & 1.8~(0.0)/5 & 0 (0) & 0.6~(0.0)/5 & \textbf{25 (3)} & 0 (0) & 0.3~(0.0)/5  \\
   DMVAE (36-D) & \textbf{1 (0)} & 1.8~(0.0)/5 & 0 (0) & 0.6~(0.1)/5 & 14 (0) & 0 (0) & 0.5~(0.1)/5 \\
   MoPoE (24-D) &  0 (0) & 1.8~(0.0)/5 & 0 (0) & \textbf{0.7~(0.0)/5} & 17 (1) & 0 (0) & \textbf{1.0~(0.0)/5}  \\
    \toprule
   MMVAE (32-D) & 0 (0) & 1.8~(0.0)/5 & 0 (0) & 0.4~(0.1)/5 & 15 (0) & 0 (0) & 0.5~(0.4)/5  \\
   MVAE (46-D) &  0 (0) & 1.8~(0.0)/5 & 0 (0) & 0.6~(0.1)/5 & \textbf{24 (2)} & 0 (0) & 0.6~(0.1)/5 \\
   DMVAE (32-D) & 0 (0) & 1.8~(0.0)/5 & 0 (0) & 0.4~(0.1)/5 & 14 (1) & 0 (0) & 0.4~(0.1)/5 \\
   MoPoE (32-D) & 0 (0) & 1.8~(0.0)/5 & 0 (0) & \textbf{0.7~(0.3)/5} & 17 (2) & 0 (0) & \textbf{1.1~(0.1)/5}   \\
    
    \bottomrule
  \end{tabularx}
\end{table}

\begin{figure}[!htp]
\centering
\includegraphics[width=1\textwidth]{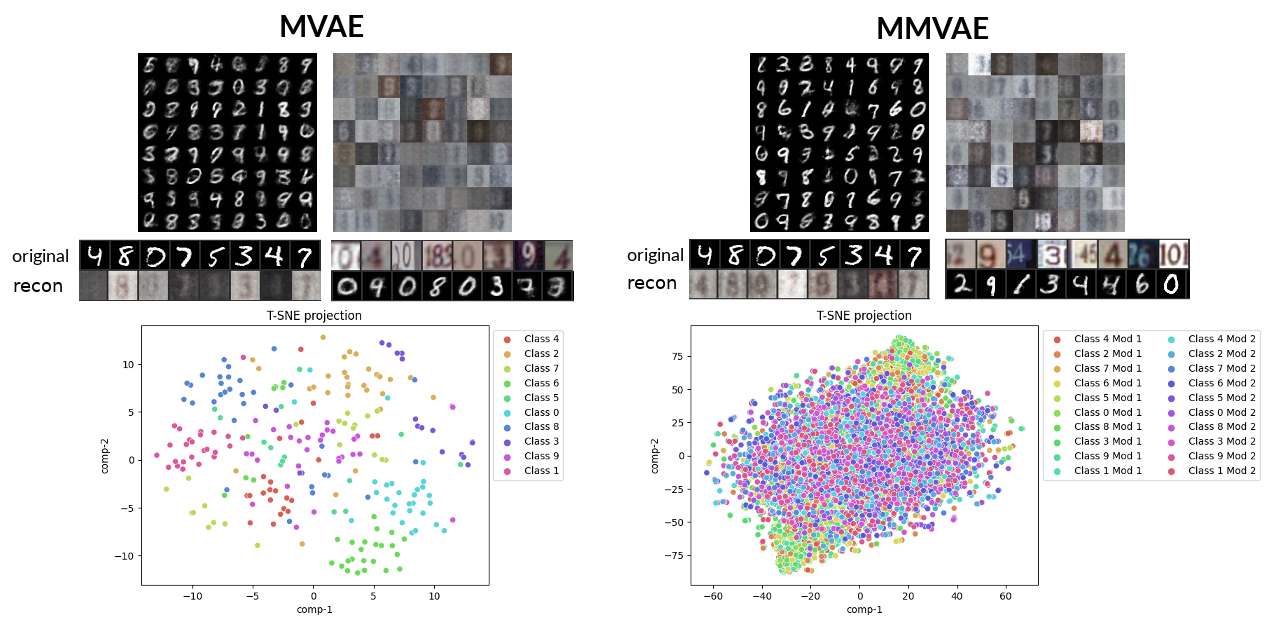}
\caption{Results for the MVAE and MMVAE models trained on the MNIST-SVHN dataset using our toolkit. For MMVAE, we used the DREG objective as proposed by the authors, MVAE was trained with ELBO. We used the encoder and decoder networks from the original implementations. The top figures are traversals for each modality, below we show cross-generated samples. The bottom figures are T-SNE visualizations of the latent space - please note that for MVAE we show samples from the single joint posterior, while for MMVAE we show samples for both modality-specific distributions. }
\label{fig:mnistsvhn}
\end{figure}

\begin{figure}[!htp]
\centering
\includegraphics[width=1\textwidth]{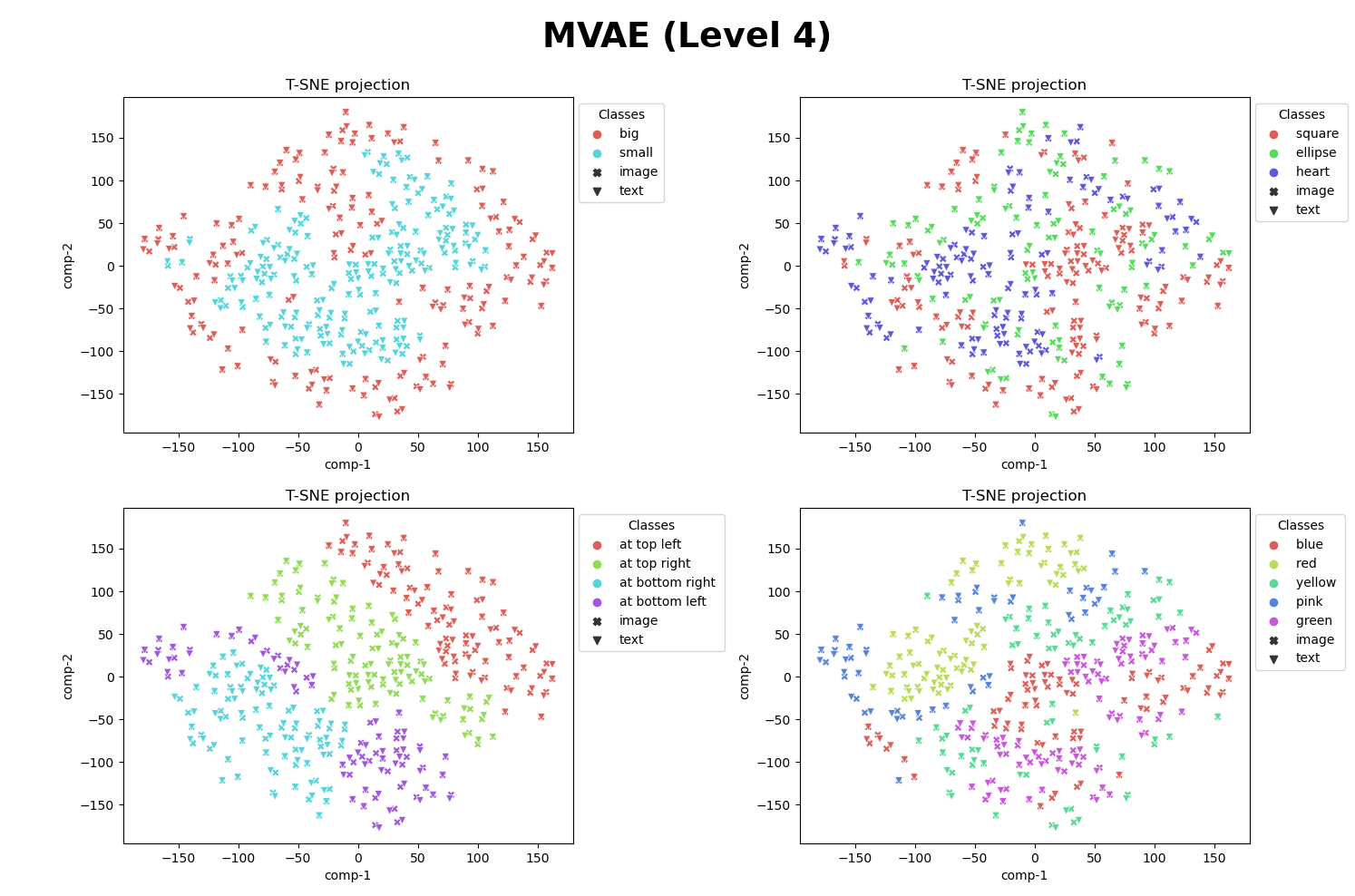}
\caption{T-SNE visualizations for the MVAE model's (16-D) joint latent space trained on \dataset Level 4. We show the latent space for each of the 4 features (size, shape, position and colour) individually.}
\label{fig:mvaetsne}
\end{figure}

\begin{figure}[!htp]
\centering
\includegraphics[width=1\textwidth]{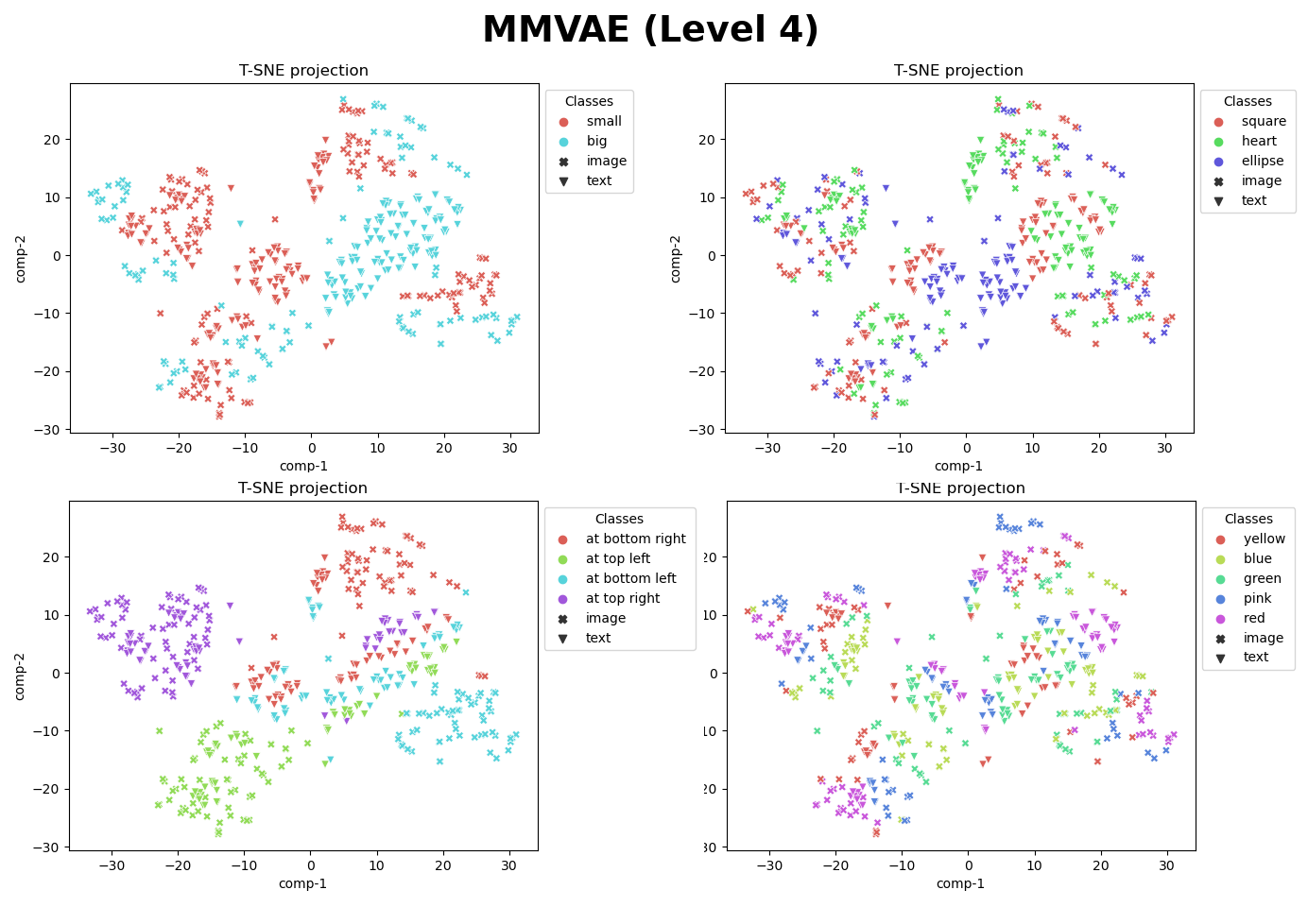}
\caption{T-SNE visualizations for the MMVAE model's (24-D) unimodal latent spaces trained on \dataset level 4. We show the latent space for each of the 4 features (size, shape, position and colour) individually.}
\label{fig:mmvaetsne}
\end{figure}

\begin{figure}[!htp]
\centering
\includegraphics[width=1\textwidth]{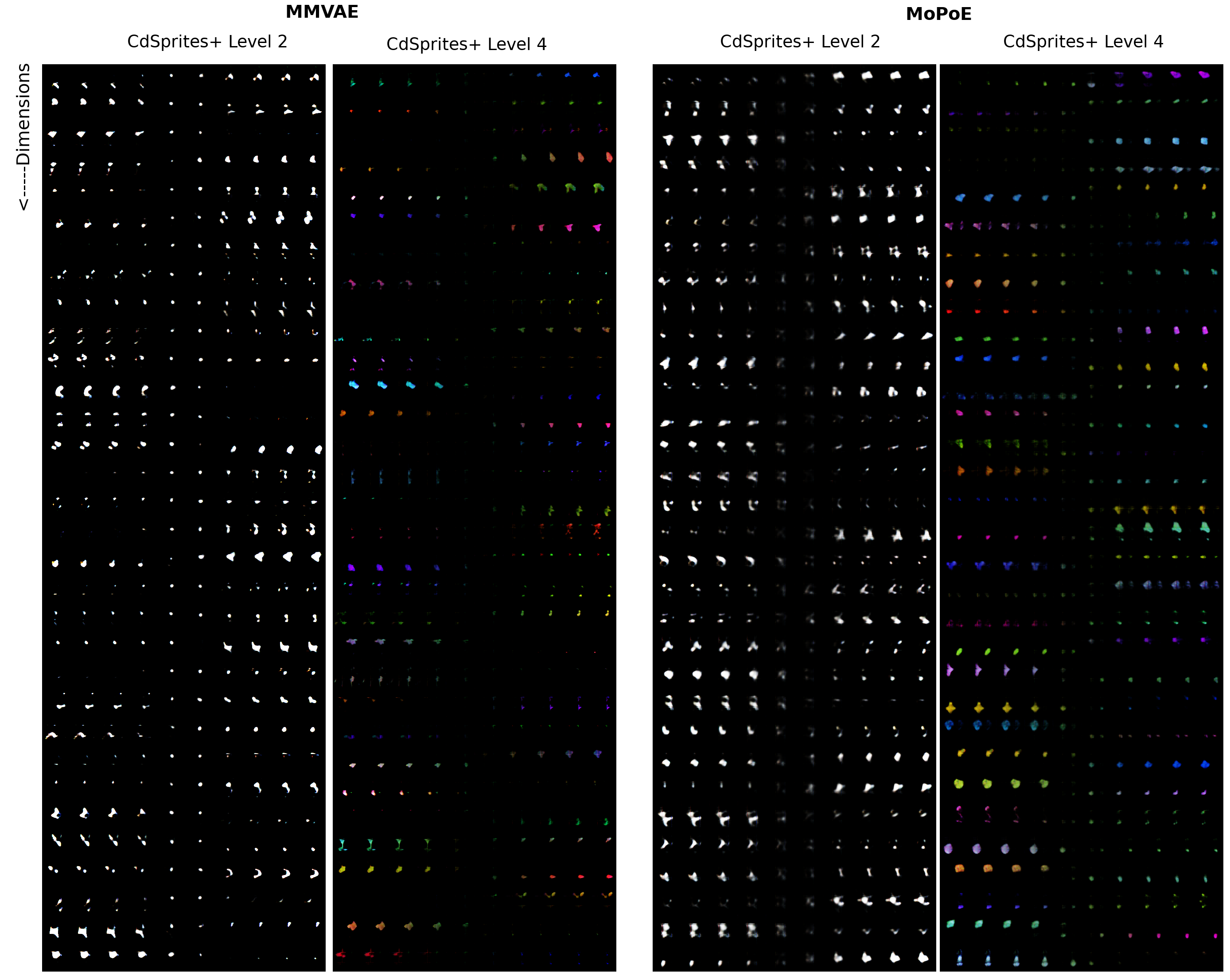}
\caption{\added{Image traversals for the MMVAE and MoPoE models for the CdSprites+ Levels 2 and 4. Each row is one out of 32 dimensions of the latent space, each column is the single sampled vector from the traversal range (-6,6).}}
\label{fig:imgtraversals}
\end{figure}

\begin{figure}[!htp]
\centering
\includegraphics[width=1\textwidth]{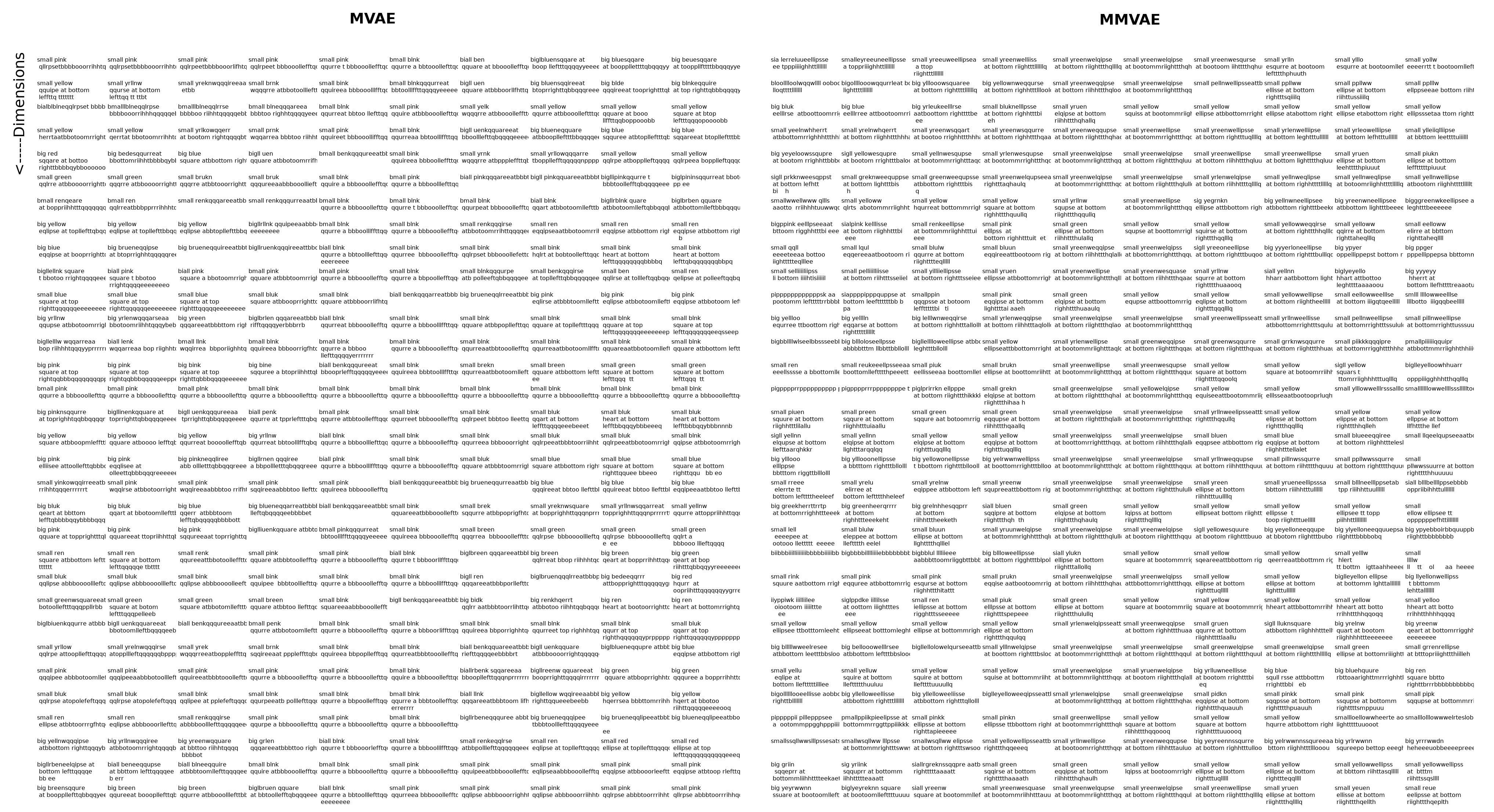}
\caption{\added{Text traversals for the MVAE and MMVAE models for the CdSprites+ Level 4. Each row is one out of 32 dimensions of the latent space, each column is the single sampled vector from the traversal range (-6,6). Note that we did not set the desired length of the text output, the model thus always generated the maximum number of characters.}}
\label{fig:txttraversals}
\end{figure}

\subsubsection{Detailed Results}
\label{sec:resultsdetailed}
In Tables \ref{tab:level1}, \ref{tab:level2}, \ref{tab:level3}, \ref{tab:level4} and \ref{tab:level5}, we show the comparison of the MVAE, MMVAE, DMVAE and MoPoE models on the 5 difficulty levels of the \dataset dataset. Here we varied the latent dimensionality (16-D to 32-D) with the fixed batch size of 32. The values are the mean cross-generation and joint-generation accuracies over 5 seeds with the standard deviations listed in brackets. According to the  \textit{Strict} metrics (with zero noise tolerance, see Sec. \ref{ssec:eval}), all models failed in both tasks at Levels 4 and 5. The  \textit{Feature} and \textit{Letter} accuracies significantly decrease across levels as the complexity increases. You can see the T-SNE visualizations for the MVAE and MMVAE models trained on Level 4 in Figs. \ref{fig:mvaetsne} and \ref{fig:mmvaetsne} .

\subsection{\mycomment{Verifying correctness of model implementation}}
\label{sec:reproduce}
\mycomment{To verify the correctness of our implementation for each model, we have reproduced selected experiments from the original papers using our toolkit. We provide both the original and our results below.}

\vspace{10mm}
\subsubsection{\mycomment{MMVAE}}
\mycomment{To verify that our implementation of the MMVAE [7] model is correct, we reproduced the experiments using the MNIST-SVHN dataset. We used the same model configuration and parameters as in the original report, i.e. the Mixture-of-Experts mixing with the DREG training objective, latent size 20 and 30 samples drawn from the joint posterior and the likelihood scaling for each modality was adjusted according to the varying dimensionalities. We used the same encoder and decoder architectures as in the original paper. After training, we calculated the joint- and cross-coherencies using the adapted original evaluation script (please see the original paper for the evaluation details [7]). The results are shown in Table \ref{tab:mmvaerepro}, the config files for reproducing the experiment are also provided on our GitHub. Please note that the results in Table \ref{tab:mmvaerepro} are different from those depicted in the main paper, Table 3. This is because here we unified the training hyperparameters with the original paper setup. However, we found that setting the likelihood scaling to 1 for both modalities produces more balanced results (in terms of MNIST/SVHN accuracies) and used thus this setup for the comparative study.}

\subsubsection{MVAE}
In the original MVAE paper [5], the results are reported in terms of marginal log-likelihoods. We reproduced the FashionMNIST experiment with a 64-D latent space, batch size 100, and likelihood scaling of 10 for the labels and 1 for the images, as reported in the public code. The results can be seen in Table \ref{tab:mvaerepro}.

\begin{table}[t]
  \caption{Reproduced MNIST-SVHN results for the MMVAE model with our multimodal VAE toolkit. We show the digit classification accuracies (\%) of latent variables (\textit{MNIST}) and (\textit{SVHN}), and the probability of digit matching (\%) for  cross- and joint-generation. For our results, we also show in brackets the variance of the results calculated over 3 seeds.}
  \label{tab:mmvaerepro}
  \centering
 \begin{tabularx}{\textwidth}{@{}p{2.8cm}p{1.6cm}p{1.6cm}p{2.2cm}p{2.2cm}p{1.9cm}@{}}
    \toprule
    Version & MNIST & SVHN & MNIST $\rightarrow$ SVHN & SVHN $\rightarrow$ MNIST & Joint  \\
    \midrule
     Original & 91.3 & 68.0 & 86.4 &  69.1 & 42.1  \\
     Reproduced (Ours) & 87.6 (5.2) & 70.4 (4.6) &  82.7 (5.2) & 72.5 (4.9) & 45.3 (3.1) \\
    \bottomrule
  \end{tabularx}
 \vspace{2mm} 
 
% \vspace{-4mm} 
%\end{table}
%\begin{table}[!htp]
  \caption{\mycomment{\mycomment{Reproduced FashionMNIST results for the MVAE model with our multimodal VAE toolkit. We show the estimated marginal log-likelihoods (lower is better). For our results, we also show in brackets the variance of the results calculated over 3 seeds.}}}
  \label{tab:mvaerepro}
  \centering
 \begin{tabularx}{\textwidth}{@{}p{3.5cm}p{3cm}p{3cm}p{3cm}@{}}
    \toprule
    Version & $log{p(x_1)}$ & $log{p(x_1, x_2)}$  & $log{p(x_1|x_2)}$ \\
    \midrule
     Original &  -232.535 & -233.007 & -230.695  \\
     Reproduced (Ours) & -234.15 (1.52) & -233.89 (2.61) & -232.56 (3.12)  \\
    \bottomrule
  \end{tabularx}
 \vspace{2mm} 
%\end{table}

%\begin{table}[!htp]
  \caption{Reproduced PolyMNIST results for the MoPoE model with our multimodal VAE toolkit. We show the Coherence Accuracy (\%) of conditionally generated samples (excluding the input
modality) (\textit{1 Mod}, \textit{2 Mods}, \textit{3 Mods} and \textit{4 Mods} stand for the number of input modalities) and the joint coherence (\textit{Joint}). For our results, we also show in brackets the variance of the results calculated over 3 seeds.}
  \label{tab:mopoerepro}
  \centering
 \begin{tabularx}{\textwidth}{@{}p{3.3cm}p{1.7cm}p{1.7cm}p{1.7cm}p{1.7cm}p{1.7cm}@{}}
    \toprule
    Version & 1 Mod &  2 Mods & 3 Mods & 4 Mods & Joint \\
    \midrule
     Original & 67 & 78 & 80 &  83 & 12 \\
     Reproduced (Ours) & 66 (4) & 73 (5) & 81 (3) & 82 (5) & 11 (3)   \\
    \bottomrule
  \end{tabularx}
 \vspace{2mm} 
%\end{table}

%\begin{table}[!htp]
  \caption{\mycomment{Reproduced MNIST-SVHN results for the DMVAE model with our multimodal VAE toolkit. We show the probability of digit matching (\%) for  cross- and joint-generation. For our results, we also show in brackets the variance of the results calculated over 3 seeds.}}
  \label{tab:dmvaerepro}
  \centering
\begin{tabularx}{\textwidth}{@{}p{3.cm}p{3cm}p{3cm}p{3cm}@{}}
    \toprule
    Version & MNIST $\rightarrow$ SVHN & SVHN $\rightarrow$ MNIST & Joint  \\
    \midrule
     Original &  88.1 & 83.7 & 44.7  \\
     Reproduced (Ours) & 84.5 (4.7) & 82.2 (3.1) & 44.9 (3.6)  \\
    \bottomrule
  \end{tabularx}
 %\vspace{-4mm} 
\end{table}

\begin{figure}
  \centering
  \includegraphics[width=0.5\linewidth]{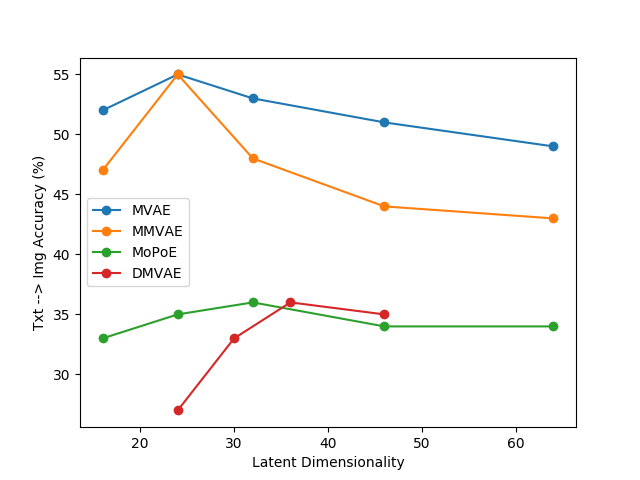}\hfill
  \includegraphics[width=0.5\linewidth]{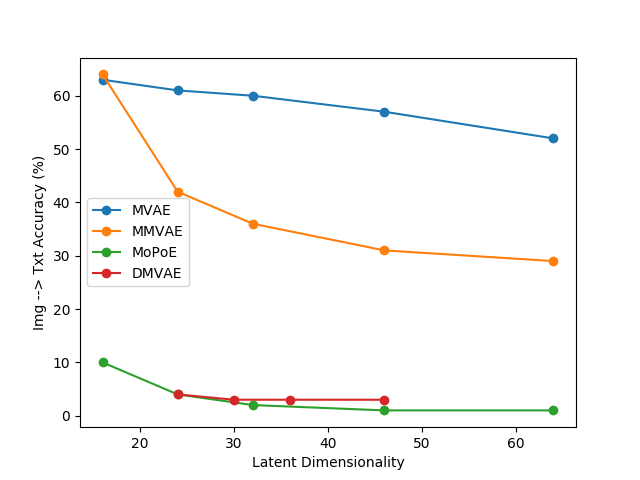}
  \caption{Example of an automated visualization generated by our toolkit. We show accuracies for the Txt $\rightarrow$ Img (left) and Img $\rightarrow$ Txt (right) cross-generations for CdSprites+ Level 1 dataset based on the used latent dimensionality. We compare the 4 implemented models. Note that for DMVAE, we trained with different dimensionalities as there are private and shared latent variables.}
  \label{fig:latentdim}
\end{figure}

\subsubsection{\mycomment{MoPoE}}
\mycomment{For verification that the MoPoE model is correct, the reproduction was performed on the PolyMNIST dataset. Based on the original implementation, we used the 512-D latent space, Laplace prior distributions, and $\beta=2.5$. After training, we calculated the cross-coherencies conditioned on 1, 2, 3 or 4 modalities as reported in the paper. The results are shown in Table \ref{tab:mopoerepro}.}

\subsubsection{\mycomment{DMVAE}}
\mycomment{We reproduced the MNIST-SVHN experiment for the DMVAE model. The reproduced model configuration included shared latent dimensionality $Dim_{shared}=10$, the private latent dimensionalities were $Dim_{MNIST}=1$ and for $Dim_{SVHN}=4$. The used $\beta$ parameter was 1, and batch size 100. We used the same encoder and decoder networks and an adapted script for calculating the cross- and joint-coherencies. The results are in Table \ref{tab:dmvaerepro}.}

\begin{table}
    \centering
    \caption{\added{Comparison of the multimodal VAE models used in our toolkit as described by \citet{suzuki2022survey}. "Aggregated inference" refers to whether the model learns joint posterior for all modalities, “Modality-specific” means whether the model
also learns modality-specific (private) latent variables and “Scalable” refers to whether the computational costs grow exponentially with the number of modalities.}}
\label{tab:modelcompar}
    \begin{tabular}{lccc}
        \toprule
        Model & Aggregated Inference & Modality-specific & Scalable \\
        \midrule
        MVAE \citep{wu2018multimodal} & \cmark & \xmark & \cmark \\
        MMVAE \citep{shi2019variational} & \xmark & \xmark & \cmark \\
        MoPoE \citep{sutter2021generalized} & \cmark & \xmark & \xmark \\
        DMVAE \citep{lee2021private}& \cmark & \cmark & \cmark \\
        \bottomrule
    \end{tabular}
\end{table}

\end{document}